\documentclass[pdflatex,sn-nature,iicol]{sn-jnl}

\usepackage{graphicx}
\usepackage{booktabs}
\usepackage{amsmath,amssymb}
\usepackage{xcolor}
\usepackage{colortbl}
\usepackage{multirow}
\usepackage{array}
\usepackage{pifont}
\graphicspath{{./figures/}}
\definecolor{tabgroup}{RGB}{243,243,243}
\definecolor{tabbase}{RGB}{240,240,240}
\definecolor{tabours}{RGB}{222,235,247}
\newcommand{\cmark}{\ding{51}}
\newcommand{\xmark}{\ding{55}}
\newcommand{\lowerbetter}{$\downarrow$}
\newcommand{\grouprow}[2]{\rowcolor{tabgroup}\multicolumn{#1}{l}{\textit{#2}}\\}
\newcommand{\tablestyle}{\small\setlength{\tabcolsep}{4pt}\setlength{\aboverulesep}{0pt}\setlength{\belowrulesep}{0pt}\renewcommand{\arraystretch}{1.2}}
\newcommand{\SI}[1]{Supplementary #1}
\newcounter{snote}
\newcommand{\snote}[2]{\refstepcounter{snote}\section*{Supplementary Note \thesnote: #1}\label{#2}}
\hypersetup{colorlinks=false}
\begin{document}

\title[Actively trained neural operator for near-explicit chemistry]{Emulating thousands of near-explicit chemical species with an actively trained neural operator}

\author*[1,2,3]{\fnm{Zhi-Song} \sur{Liu}}\email{zhisong.liu@lut.fi}
\author[2,3]{\fnm{Wenqing} \sur{Peng}}
\author[1,2,3]{\fnm{Ammar} \sur{Kheder}}
\author[4]{\fnm{Risto} \sur{Makkonen}}
\author[1,2,3]{\fnm{Michael} \sur{Boy}}
\affil[1]{Department of Computational Engineering, LUT University, Finland} \affil[2]{Atmospheric Modelling Center, Lahti (AMC-Lahti), Finland} \affil[3]{Institute for Atmospheric and Earth System Research (INAR), University of Helsinki, Finland} \affil[4]{Finnish Meteorological Institute (FMI), Finland}

\abstract{Earth-system models represent atmospheric chemistry with condensed mechanisms because near-explicit mechanisms such as the Master Chemical Mechanism (MCM) are too costly to solve in every grid cell. Neural emulators could remove this constraint, but most advance few species in short time steps. Here we present FNO-MoE, a Fourier neural operator with a mixture of experts that predicts 4,691 MCM species over the next 30 minutes in one call. Its experts' disagreement selects which box-model simulations to run; at a fixed budget, this active sampling lowers one-step error by 12--31\% relative to static and random sampling. After fine-tuning on six-hour trajectories, none of 14 six-hour test cases diverges, against all 14 for the one-hour model. The emulator reproduces six-hour trajectories with a median per-species $R^2$ of 0.998 and runs about 750 times faster than the solver on one CPU core. Skill drops under time-varying forcing ($R^2=0.870$), the next step towards coupling.}


\maketitle

\section*{Introduction}

Gas-phase oxidation sets the lifetimes of methane, ozone and other short-lived climate forcers and air pollutants, and its OH and ozone budgets define the oxidizing capacity of the troposphere. The same chemistry converts volatile organic compounds into highly oxygenated, low-volatility products that form secondary organic aerosol and drive the formation and growth of new particles towards cloud condensation nuclei \cite{ehn2014,kirkby2016}. Which products form depends on individual peroxy-radical pathways, the level of detail that condensed mechanisms lump together. Chemical transport and Earth-system models must solve this chemistry in every grid cell at every chemistry time step \cite{brasseur2017}. They can afford to do so only with condensed mechanisms that lump organic compounds into tens to hundreds of surrogate species, and even then gas-phase chemistry typically takes 36--95\% of the runtime of a chemical transport model \cite{liu2025film}. The Master Chemical Mechanism (MCM) instead describes tropospheric oxidation near-explicitly \cite{jenkin1997,saunders2003,mcm}; the configuration used here links 6,068 species through 18,490 reactions. Its stiff ordinary differential equations are expensive to integrate \cite{kpp}, so near-explicit chemistry is used mainly in box and column models \cite{arca}, and large-scale models cannot follow the individual oxidation intermediates and condensable products that the MCM resolves.

A neural emulator could make this detail affordable, but only if it fits the setting in which it will run. Inside a host model the chemistry routine never starts from a box-model initial condition. It receives whatever species vector transport, emissions and the previous step have left, together with the local physical conditions, and must return the state at the next coupling time. Such an emulator should therefore meet three requirements. It should (i) map the full chemical state and the physical conditions to future states from any point along a trajectory; (ii) need as few calls as possible per simulated hour, because every call is repeated in every grid cell; and (iii) remain accurate when its own predictions are fed back over many hours. We call such chained integration a rollout; keeping it stable is a known difficulty for autoregressive neural solvers \cite{brandstetter}.

Neural networks have become credible surrogates across Earth-system modelling, from subgrid parameterization \cite{rasp2018} to global weather and climate simulation \cite{graphcast,pangu,neuralgcm}. For gas-phase chemistry, learned solvers have been built with random forests \cite{keller2019}, autoencoders \cite{kelp2020}, mass- and energy-conserving networks \cite{sturm2020}, attention-based operators \cite{liu_chemnne} and condition-modulated residual networks \cite{liu2025film}. Two lines of work also address long-run stability: online training kept a neural solver for a 12-species mechanism stable in year-long global simulations \cite{kelp2022}, and SIMADy learns a sparse mass-action network on six latent species of the MCM whose trajectories are bounded by construction \cite{simady}. Together, these studies establish that learned chemistry can be fast and stable. Two limitations nevertheless separate them from the requirements above.

First, most emulators learn a step-wise solver: a network that advances the state by one chemistry time step, or a reduced system of differential equations that must itself be integrated \cite{simady}. The cost is paid at every step, so the attainable speedup is capped by the number of calls. A step-wise emulator that is 80--130 times faster than the numerical solver offline, for example, accelerates the chemistry module of a chemical transport model by only 2--4.6 times once embedded \cite{liu2025film}. Second, most emulators predict a small state, either condensed mechanisms of 12 to 72 species \cite{kelp2022,liu2025film} or a few target species, such as ozone through six latent variables \cite{simady} or five trace gases from an air-quality foundation model that does not enforce chemical consistency among them \cite{hu2026aurora}. A small state also hides a data problem. Training simulations are drawn from predefined scenarios or at random, without asking which would be most informative. At near-explicit scale this becomes a binding constraint: here the initial condition of a reference simulation has 150 dimensions, so any affordable budget covers this space only sparsely, and more representative sampling has been named a priority for chemistry emulators \cite{liu2025film}.

\begin{figure*}[!t]
\centering
\includegraphics[width=\textwidth]{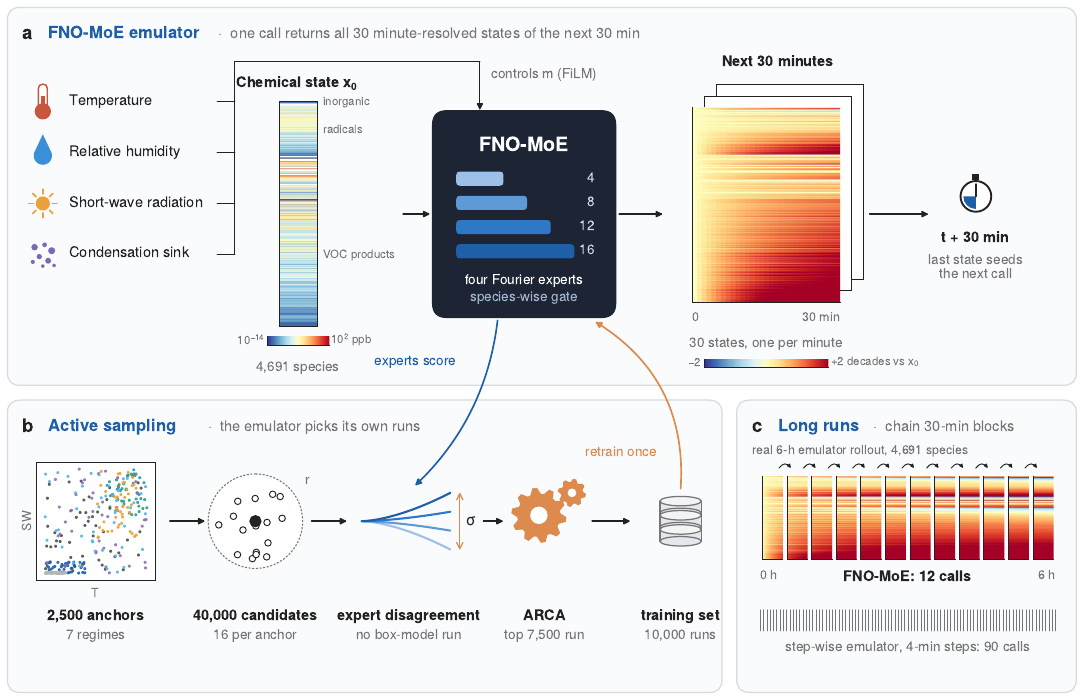}
\caption{\textbf{One network both emulates near-explicit chemistry and chooses its own training simulations.} \textbf{a}, The FNO-MoE emulator maps the full retained chemical state $x_0$ (4,691 species, shown for a real ARCA state, grouped by chemical family and coloured by mixing ratio from $10^{-14}$ (blue) to $10^{2}$\,ppb (red)) and four physical controls $m$ (temperature, relative humidity, short-wave radiation and condensation sink) to all 30 minute-resolved states of the next 30 minutes in one forward pass (stacked sheets: the emulator's prediction of the change from $x_0$ for the same case; blue, decrease; red, increase). Inside, the state is lifted and broadcast over the horizon with a time coordinate; four independent Fourier neural operator experts retain 4, 8, 12 and 16 Fourier modes, each applying four spectral blocks (Fourier transform $F$, weights $R_k$ on the lowest $k$ modes, inverse transform $F^{-1}$, a $1\times1$ bypass $W$, FiLM($m$) and GELU); a static species-wise gate $\pi=\mathrm{softmax}(g)$ combines the experts' increments as $\sum_e\pi_{s,e}\Delta^{(e)}$, which is added to $x_0$. The last predicted state seeds the next call. \textbf{b}, Single-round active sampling: after a warm-up of 2,500 ARCA simulations, one near each of 2,500 anchor states drawn from seven atmospheric regimes (shown: temperature versus short-wave radiation), 16 Sobol candidates are drawn within $r=0.10$ of each anchor (40,000 in total); the frozen experts score each candidate by the spread $\sigma$ of their predictions for its first 30 minutes, without running the box model; the three highest-scoring candidates per anchor (7,500) are simulated by ARCA, and the emulator is retrained once on all 10,000 simulations. Blue arrows and glyphs, emulator only; orange, ARCA box-model runs. \textbf{c}, A real six-hour rollout of the deployed emulator drawn as its twelve chained 30-min calls (rows, species; colour, change from the starting state on the same blue--red scale as in a), against the 90 calls of a step-wise emulator with a 4-min chemistry step.}\label{fig:overview}
\end{figure*}

Here we address both limitations with FNO-MoE, a Fourier neural operator \cite{fno} with a mixture of spectral experts that emulates 4,691 MCM species (Fig.~\ref{fig:overview}). Instead of a step-wise solver, it learns an operator that maps the current state and physical conditions to all 30 minute-resolved states of the next 30 minutes in one forward pass, so six hours take twelve calls regardless of the internal step size of the stiff solver. To meet the data problem, the emulator chooses its own training simulations: its four experts form a built-in committee, and their disagreement ranks candidate box-model simulations before any is run. Finally, fine-tuning on windows from six-hour trajectories exposes the operator to the slowly evolving states that it reaches when its predictions are chained.

We show that, at a fixed budget, emulator-guided sampling gives the lowest one-step error in every metric and atmospheric regime; that six-hour fine-tuning removes divergence in all 14 distinct test cases; that the deployed model reproduces six-hour trajectories of all retained species and transfers to novel compositions with a modest loss of skill; and that it runs about 750 times faster than the reference on one CPU core. We also identify where it fails, under extreme and time-varying forcing, which defines what remains before coupled deployment.

\section*{Results}

\subsection*{One call advances the full chemical state by 30 minutes}

The numerical reference is the Atmospherically Relevant Chemistry and Aerosol box model (ARCA) \cite{arca} running the MCM. Each reference simulation is seeded by an initial condition vector $\theta$ of four physical controls (temperature, relative humidity, short-wave radiation and condensation sink) and 146 precursor concentrations, and is integrated for one or six hours under constant controls with the state saved every minute (Methods). The emulator never receives $\theta$. In line with requirement (i), training pairs are cut from anywhere along each trajectory: the input is the full retained state $x_0$, that is, the 4,691 species, products included, that pass a low-concentration filter, together with the four controls $m$, and the target is the 30 minute-resolved states that follow.

FNO-MoE (Fig.~\ref{fig:overview}a) broadcasts $x_0$ over the 30-min horizon, appends a time coordinate and passes the result through four independent Fourier neural operator experts, each conditioned on $m$ by feature-wise linear modulation \cite{film}. Experts with 4, 8, 12 and 16 Fourier modes resolve progressively faster temporal variation, which suits a state whose species range from short-lived radicals to slowly accumulating products. A learned species-wise gate combines the experts' increments \cite{moe}, and the result is added to $x_0$. Because the experts are independently parameterized, they also form a committee whose disagreement guides sampling.

Every reference simulation starts with the precursors at their sampled values and all other species at a concentration floor, so many products and radicals rise steeply in the first five minutes. A host model would rarely see this stiff transient, because it calls the chemistry from an evolved state. We therefore name model variants by whether training includes windows that start in the transient (Stiff) or not (NoStiff) and by the training horizon (1h, or 6h after fine-tuning on six-hour trajectories). All variants share one architecture of 7.8 million parameters.

\subsection*{The emulator's experts choose which simulations to run}

Which simulations are run decides which state transitions the emulator can learn. Uniform sampling of the 150-dimensional space spends most of the budget on combinations that rarely co-occur in the atmosphere; sampling evenly around plausible states avoids this, but still spends simulations where the emulator is already accurate. We therefore let the emulator choose, using single-round active sampling \cite{settles} (Fig.~\ref{fig:overview}b).

Seven hand-built regimes encode which meteorology and precursor mixtures co-occur: clean biogenic summer day, urban polluted, cold low-biogenic winter night, mixed background, marine remote background, biomass-burning plume and polar pristine night. Sampling within them gives 2,500 anchor states (\SI{Tables~\ref{stab:regimes} and~\ref{stab:ranges}}). After a warm-up of one simulation near each anchor, the emulator is frozen, and 16 fresh Sobol candidates are drawn within a small neighbourhood of every anchor (radius 0.10 in normalized units). Each candidate is scored by the disagreement among the four experts, averaged over species and minutes, and only the three highest-scoring candidates per anchor are simulated. The emulator is then retrained once on all 10,000 simulations; in total, 40,000 candidates are scored and 7,500 simulated. Scoring is cheap, about 16\,ms per candidate on one CPU core against minutes for a box-model simulation, and the fixed quota per anchor guarantees that every regime receives new data.

\begin{table*}[!t]
\caption{\textbf{Active local sampling gives the lowest one-step error in every metric.} Normalized-log RMSE on the 2,500 regime anchors for three strategies with identical architecture and a matched budget of 10,000 one-hour simulations. Initial windows start in the first five minutes, during the stiff initialization transient; post-transient windows start at or after minute five. All species: each dataset's retained species (4,691 for active, 4,713 for static and 4,725 for random sampling); Key: 204 key species (\SI{Note~\ref{snote:key}}). Lower is better; best values in bold.}\label{tab:sampling}
\centering
\tablestyle
\begin{tabular}{lcccc}
\toprule
\multirow{2}{*}{\textbf{Sampling strategy}} & \multicolumn{2}{c}{\textbf{All species}} & \multicolumn{2}{c}{\textbf{Key species} (204)} \\
\cmidrule(lr){2-3}\cmidrule(l){4-5}
 & Initial\lowerbetter & Post-transient\lowerbetter & Initial\lowerbetter & Post-transient\lowerbetter \\
\midrule
\rowcolor{tabbase} Whole-space random & 0.2670 & 0.0108 & 0.1315 & 0.0063 \\
\rowcolor{tabbase} Static local & 0.1469 & 0.0085 & 0.0716 & 0.0063 \\
\rowcolor{tabours} \textbf{Active local (this work)} & \textbf{0.1289} & \textbf{0.0075} & \textbf{0.0641} & \textbf{0.0051} \\
\bottomrule
\end{tabular}
\end{table*}
\begin{figure*}[!t]
\centering
\includegraphics[width=\textwidth]{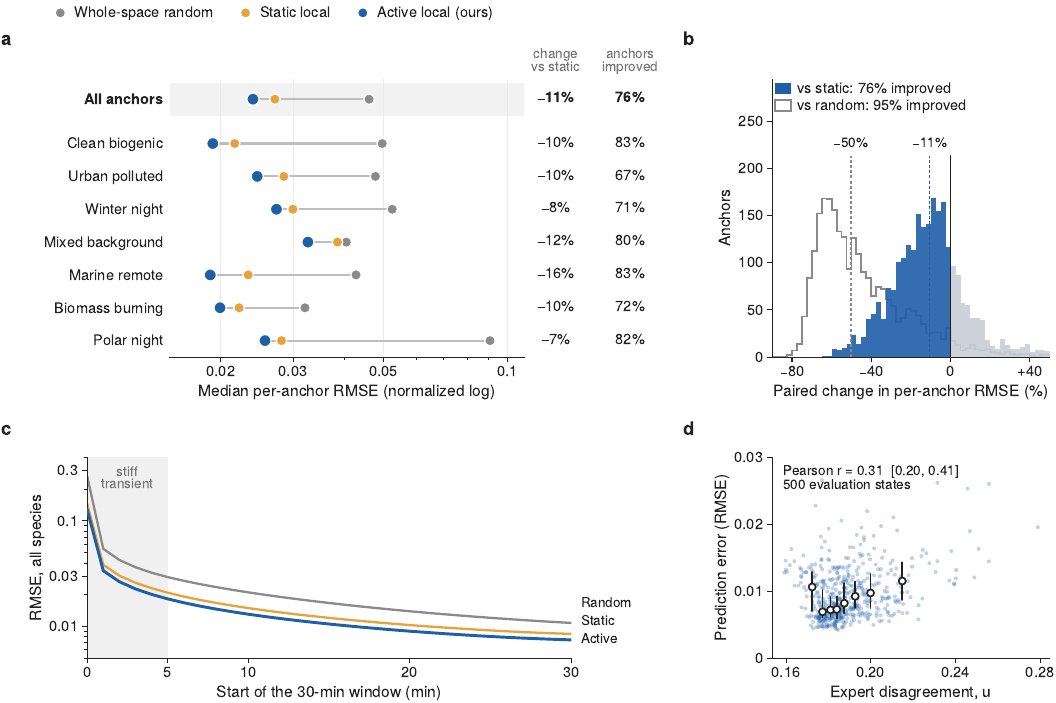}
\caption{\textbf{Emulator-guided sampling lowers one-step error on most anchors at a fixed budget of 10,000 simulations.} Grey, whole-space random; amber, static local; blue, active local sampling (this work). \textbf{a}, Median per-anchor all-species RMSE of the three strategies for all 2,500 evaluation anchors and for each atmospheric regime; right, the median paired change of active relative to static sampling and the share of anchors on which active sampling is more accurate. All regime-level changes have 95\% paired-bootstrap intervals that exclude zero (all anchors: $-10.5\%$ [$-11.4$, $-10.0$]; improved on 76\% [74, 78] of anchors). \textbf{b}, Distribution of the paired per-anchor change in RMSE of active relative to static sampling (filled; blue, improved; grey, worse) and relative to random sampling (outline); dashed lines, medians ($-11\%$ and $-50\%$). \textbf{c}, RMSE of the 30-min block versus its starting offset within the reference simulation; shading marks the floor-seeded transient (first 5\,min), where errors are 17--25 times larger than later. \textbf{d}, Pre-gate expert disagreement versus prediction error for 500 evaluation states (Pearson $r=0.31$, 95\% bootstrap interval $[0.20,0.41]$); circles and bars, median and interquartile range of the error in eight equal-count bins of disagreement. Disagreement ranks candidates usefully but is not a calibrated error estimate.}\label{fig:sampling}
\end{figure*}

The score assumes that disagreement is informative about error, and this holds moderately well. Across 500 evaluation states, expert disagreement and prediction error are positively associated (Pearson $r=0.31$, 95\% bootstrap confidence interval $[0.20,0.41]$; Fig.~\ref{fig:sampling}d), and more strongly across species ($r=0.46$, $[0.43,0.49]$). Disagreement is therefore a useful heuristic for ranking candidates, in the spirit of query by committee \cite{qbc}, but not a calibrated uncertainty estimate.

\subsection*{Active sampling lowers error at a fixed simulation budget}

We compared three strategies with the same architecture, 10,000 one-hour simulations each and the same 2,500 evaluation anchors (Methods): whole-space random sampling, static local sampling around the anchors, and active local sampling, which differs from static local sampling only in which candidates are kept. We report the root-mean-square error (RMSE) of the normalized logarithm of concentration (Eq.~\eqref{eq:norm}) for all species and for 204 key species, separately for windows that start inside and after the stiff transient. An RMSE of 0.01 corresponds to 0.04 in $\log_{10}$ concentration, a concentration error of about 10\%.

Active local sampling has the lowest error in all four metrics (Table~\ref{tab:sampling}; \SI{Fig.~\ref{sfig:errdist}}). Its post-transient all-species error, 0.0075, is 12\% lower than that of static local sampling (0.0085) and 31\% lower than that of random sampling (0.0108). Anchor locality accounts for most of the gain over random sampling in three of the four metrics, and active selection adds a further reduction. For key species after the transient, the whole gain comes from active selection (0.0063 to 0.0051).

The gain is not carried by a few anchors. Comparing the three models anchor by anchor, active sampling has the lower all-species error than static local sampling on 76\% of the 2,500 anchors (95\% bootstrap confidence interval 74--78\%) and than whole-space random sampling on 95\% (94--96\%; Fig.~\ref{fig:sampling}a,b). The median paired change in per-anchor error is $-10.5\%$ ($-11.4$ to $-10.0\%$) relative to static and $-50.3\%$ ($-51.5$ to $-49.1\%$) relative to random sampling, and every regime-level interval excludes zero. Active sampling also wins in every regime, with median changes relative to static sampling between $-6.9\%$ and $-16.4\%$, including the biomass-burning and polar-night regimes that carry the smallest design weights (Fig.~\ref{fig:sampling}a). A preliminary single-chain Metropolis variant, which visited only about 30 of the 2,500 anchors per round, performed worse than both static sampling and the adopted scheme (\SI{Table~\ref{stab:acq}}). Because the evaluation anchors also guide acquisition, this comparison tests prediction near a known design; generalization beyond it is tested below.

The same comparison shows where one-step error comes from. For every strategy, it is concentrated in the stiff transient: windows that start in the first five minutes have all-species errors 17--25 times larger than later windows (Table~\ref{tab:sampling}, Fig.~\ref{fig:sampling}c). Training without these windows (1h-NoStiff) trades full-trajectory accuracy for post-transient accuracy relative to 1h-Stiff: its full-trajectory RMSE is 0.2553 against 0.0271, and its post-transient RMSE 0.0083 against 0.0111 (values from a per-species pipeline that is not directly comparable with Table~\ref{tab:sampling}; Methods). Within one-hour simulations, the emulator reproduces the principal precursor decay at clean-biogenic, urban and biomass-burning anchors, with visible deviations in the curvature of O$_3$ (\SI{Fig.~\ref{sfig:odefits}}).

\subsection*{Six-hour fine-tuning prevents rollout divergence}

Requirement (iii) asks whether accurate 30-min blocks can be chained. Each rollout starts from the reference state at minute five and runs for six hours, with the last predicted state of each block seeding the next call. We evaluated six in-envelope cases, regime anchors withheld from training, and eight envelope-edge cases with the physical controls at the edges of the training ranges, one control at a time or all four together (Methods; \SI{Table~\ref{stab:edge}}). Besides the final RMSE we report the elemental drift, the mean departure of predicted C, H, O, N and S totals from the reference.

\begin{table*}[!t]
\caption{\textbf{Six-hour fine-tuning removes divergence and lowers final error by 88--94\%.} Six-hour autoregressive rollouts from post-transient states (minute five). RMSE (normalized log) and elemental drift at the last minute; Div., number of diverged cases (any normalized-log value $|x|>3$ or non-finite). 6-h FT, fine-tuned on six-hour trajectories; Transient, stiff initialization windows included in training. Envelope-edge RMSE and divergence are computed over the eight distinct cases; drift is averaged over the 29 runs, in which seven edge cases appear four times (Methods). Lower is better; best values in bold; the deployed model is shaded.}\label{tab:rollout}
\centering
\tablestyle\footnotesize\setlength{\tabcolsep}{3.5pt}
\begin{tabular}{lcccccccc}
\toprule
\multirow{2}{*}{\textbf{Model}} & \multirow{2}{*}{\textbf{6-h FT}} & \multirow{2}{*}{\textbf{Transient}} & \multicolumn{3}{c}{\textbf{In envelope} (6 cases)} & \multicolumn{3}{c}{\textbf{Envelope edge} (8 cases)} \\
\cmidrule(lr){4-6}\cmidrule(l){7-9}
 & & & RMSE\lowerbetter & Drift\lowerbetter & Div.\lowerbetter & RMSE\lowerbetter & Drift\lowerbetter & Div.\lowerbetter \\
\midrule
\rowcolor{tabbase} 1h-Stiff & \xmark & \cmark & 0.322 & 0.628 & 6/6 & 0.354 & 0.483 & 7/8 \\
\rowcolor{tabbase} 1h-NoStiff & \xmark & \xmark & 0.234 & 0.270 & 6/6 & 0.257 & 0.231 & 8/8 \\
\rowcolor{tabbase} 6h-Stiff & \cmark & \cmark & 0.0371 & 0.0057 & \textbf{0/6} & 0.0269 & 0.0121 & \textbf{0/8} \\
\rowcolor{tabours} \textbf{6h-NoStiff (deployed)} & \cmark & \xmark & \textbf{0.0274} & \textbf{0.0051} & \textbf{0/6} & \textbf{0.0161} & \textbf{0.0052} & \textbf{0/8} \\
\bottomrule
\end{tabular}
\end{table*}
\begin{figure*}[!t]
\centering
\includegraphics[width=\textwidth]{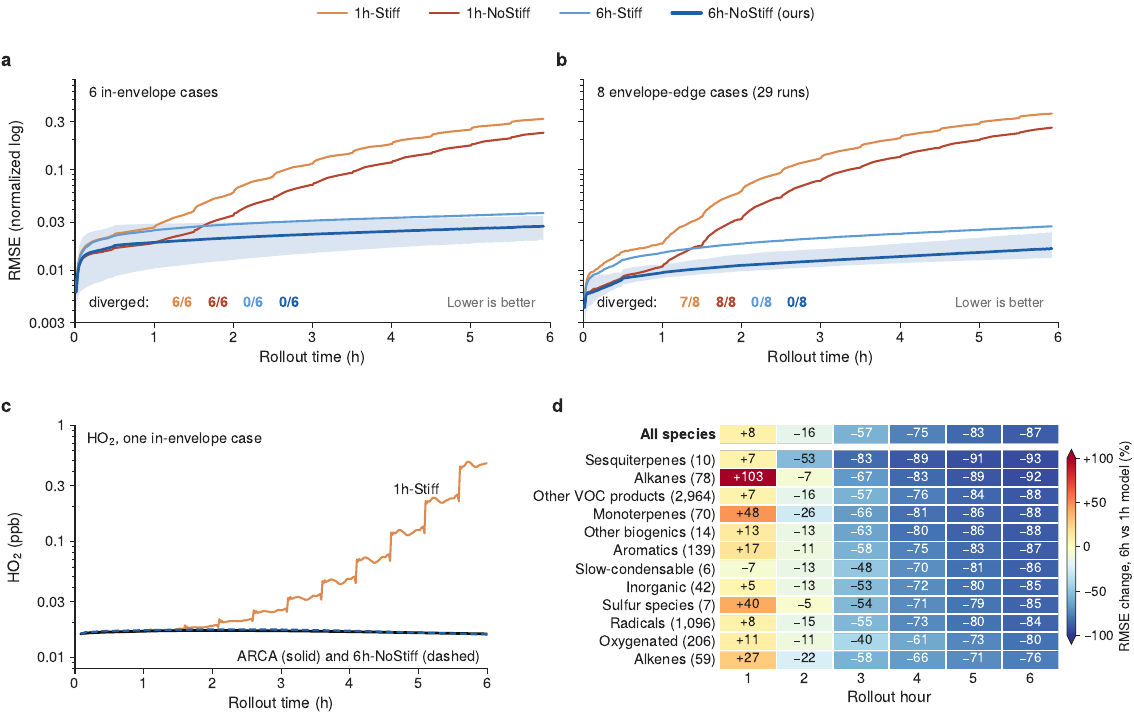}
\caption{\textbf{Six-hour fine-tuning keeps chained 30-min predictions stable.} \textbf{a},\textbf{b}, Normalized-log RMSE versus rollout time for the four variants in the six in-envelope (a) and eight envelope-edge (b) cases; lines are means over runs (b: 29 runs, eight distinct cases; Methods), shading the 10th--90th percentiles for 6h-NoStiff; coloured counts give diverged distinct cases. \textbf{c}, Hydroperoxyl radical (HO$_2$) over a six-hour rollout from a post-transient reference state: ARCA reference, 1h-Stiff and 6h-NoStiff (log axis); 1h-Stiff ends about 30 times too high. \textbf{d}, Scorecard of the RMSE of 6h-NoStiff relative to 1h-NoStiff from the same starting states, by chemical family (species counts in brackets) and rollout hour, pooled over the six in-envelope cases; blue, six-hour model more accurate; colours saturate at $\pm100\%$. The variants in c differ in both training horizon and transient inclusion; Table~\ref{tab:rollout} separates the two effects.}\label{fig:rollout}
\end{figure*}

Both one-hour models diverge in all six in-envelope rollouts (Table~\ref{tab:rollout}, Fig.~\ref{fig:rollout}). After fine-tuning on six-hour trajectories, 6h-NoStiff diverges in none; relative to 1h-NoStiff it reduces the final error from 0.234 to 0.0274 (88\%) and the elemental drift from 0.270 to 0.0051. At the envelope edges, the one-hour models diverge in seven and eight of the eight cases, whereas 6h-NoStiff diverges in none and reaches a final error of 0.0161, 94\% lower than 1h-NoStiff. The gain is consistent from case to case: the per-case final error of 6h-NoStiff has a median of 0.027 (interquartile range 0.024--0.031) in envelope and 0.015 (0.013--0.018) at the envelope edges, against medians of 0.21--0.30 for the one-hour models, and error grows gradually without runaway growth (Fig.~\ref{fig:rollout}a,b; \SI{Figs~\ref{sfig:interp}--\ref{sfig:percentiles}}). At the level of individual species, 1h-NoStiff is as accurate as 6h-NoStiff for the first two hours, after which its errors spread: the fraction of species off by more than a factor of three rises from 0.8\% at two hours to 8\% at three hours and 49\% at six hours, whereas for 6h-NoStiff it never exceeds 0.7\%. Resolved by chemical family and rollout hour, the gain is uniform across the mechanism (Fig.~\ref{fig:rollout}d): in the first hour 6h-NoStiff is slightly less accurate than 1h-NoStiff (all-species RMSE $+8\%$), but from the third hour onwards its error is 40--83\% lower in every family and 76--93\% lower in every family by the sixth hour (all species $-87\%$). An HO$_2$ trajectory illustrates the failure mode: 1h-Stiff drifts upwards at each 30-min block boundary and ends about 30 times too high, whereas 6h-NoStiff stays within 2.4\% of the reference (Fig.~\ref{fig:rollout}c).

Excluding the transient is not what provides stability. 6h-Stiff, fine-tuned from 1h-Stiff, also avoids divergence in all 14 cases, although its final errors are higher than those of 6h-NoStiff (0.0371 versus 0.0274 in envelope; 0.0269 versus 0.0161 at the envelope edges). Stability is therefore associated with the six-hour fine-tuning protocol. A plausible explanation is that one-hour training rarely shows the model the slowly evolving states reached later in a rollout, so early errors push its input outside the training distribution; because fine-tuning changes both state coverage and the amount of training, our experiments cannot separate the two. The two fine-tuned variants run at identical cost: 6h-NoStiff is more accurate when the host model evolves the chemistry smoothly between calls, whereas 6h-Stiff retains the ability to follow abrupt onsets from near-floor states, such as after emission injections (untested under coupling). We use 6h-NoStiff, the more accurate variant in both test sets, as the deployed model.

\subsection*{The full retained state is reproduced species by species}

\begin{figure*}[!t]
\centering
\includegraphics[width=\textwidth]{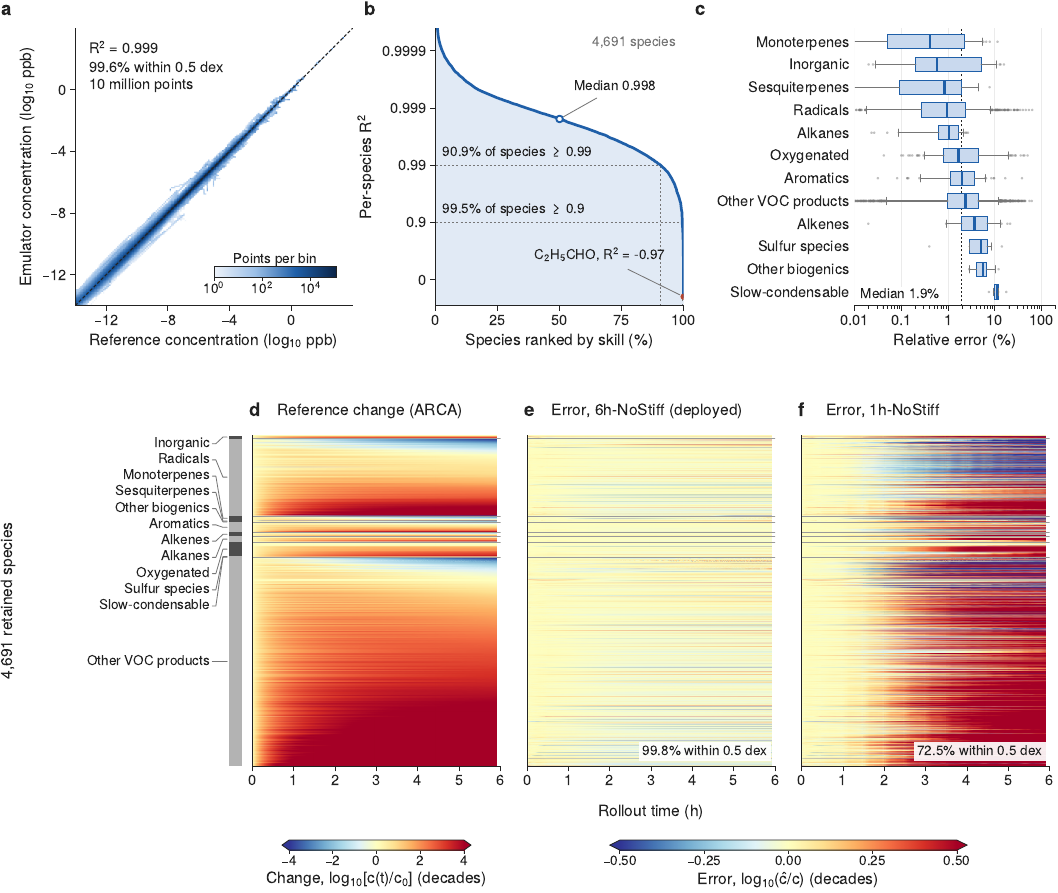}
\caption{\textbf{The deployed emulator reproduces the six-hour evolution of all 4,691 retained species.} \textbf{a}, Two-dimensional density of emulator (6h-NoStiff) versus reference concentration for all retained species at every minute of the six in-envelope rollouts ($10^7$ points; log colour scale); dashed, 1:1 line. \textbf{b}, Per-species $R^2$ (normalized-log space) of all 4,691 retained species, ranked from best to worst; the vertical axis is scaled as $-\log_{10}(1-R^2)$ so that every species is shown. Dashed guides mark the share of species with $R^2\geq0.99$ (90.9\%) and $R^2\geq0.9$ (99.5\%); open dot, median (0.998); red dot, the only species with negative $R^2$ (C$_2$H$_5$CHO, $-0.97$). \textbf{c}, Per-species relative concentration error by chemical family (median 1.9\%, dashed line). Boxes, interquartile range; line, median; whiskers, 5th--95th percentiles; dots, species beyond them. \textbf{d}, Reference change from the starting concentration, $\log_{10}[c(t)/c_0]$, for every species (rows, grouped by family) in a typical rollout (the case with the median share within 0.5 dex). \textbf{e}, Error of 6h-NoStiff, $\log_{10}(\hat c/c)$, for the same case. \textbf{f}, As e for the one-hour model 1h-NoStiff from the same starting state. In d, blue marks a decrease and red an increase; in e and f, blue marks under- and red over-prediction. Boxed numbers give the share of species--minutes within 0.5 dex in this case; over all six rollouts the shares are 99.6\% (6h-NoStiff) and 84.6\% (1h-NoStiff). Colours saturate beyond the colour-bar limits.}\label{fig:fullstate}
\end{figure*}

Aggregate errors could hide a subset of poorly predicted species, so we evaluated the deployed model species by species over the in-envelope rollouts, pooling all steps (Methods). Predicted and reference concentrations agree across more than ten orders of magnitude (Fig.~\ref{fig:fullstate}a): over $10^{7}$ species--time points, the log-concentration coefficient of determination is 0.999, and 99.6\% of predictions lie within 0.5 dex (a factor of about three) of the reference.

Agreement in level does not by itself show that the dynamics are right. Over six hours, 83--88\% of species grow, mostly oxidation products that accumulate, and more than half change by over two orders of magnitude, whereas precursors and some radicals decay (Fig.~\ref{fig:fullstate}d). The emulator reproduces this pattern species by species, with a median absolute error of 0.026 dex over all species--minutes, a concentration error of about 6\% (Fig.~\ref{fig:fullstate}e). The one-hour model, run from the same states, accumulates errors across the whole species set after two hours (Fig.~\ref{fig:fullstate}f).

Per-species scores confirm that no part of the state is neglected. Across all 4,691 retained species, the median per-species $R^2$ in normalized-log space is 0.998; 99.5\% of species reach $R^2\geq0.9$, 90.9\% reach $R^2\geq0.99$, and only one, propanal (C$_2$H$_5$CHO), has a negative $R^2$ (Fig.~\ref{fig:fullstate}b). Skill is high in every chemical family, including sesquiterpenes and slow-condensable products, which are under-represented in the mechanism. The median per-species relative error is 1.9\% of the reference concentration: below 1\% for the abundant inorganic, radical and monoterpene families, and 5--11\% for low-abundance families (sulfur species $\approx5\%$, biomass-associated biogenics $\approx6\%$, slow-condensable products $\approx11\%$), where small absolute concentrations inflate a relative metric (Fig.~\ref{fig:fullstate}c; \SI{Fig.~\ref{sfig:skill}}). The largest visible deviations in individual trajectories are in HO$_2$ and NO$_3$ and in the slow drift of SO$_2$ and CO (\SI{Fig.~\ref{sfig:traj}}).

\subsection*{Skill transfers to novel compositions but not to extreme forcing}

\begin{figure*}[!t]
\centering
\includegraphics[width=\textwidth]{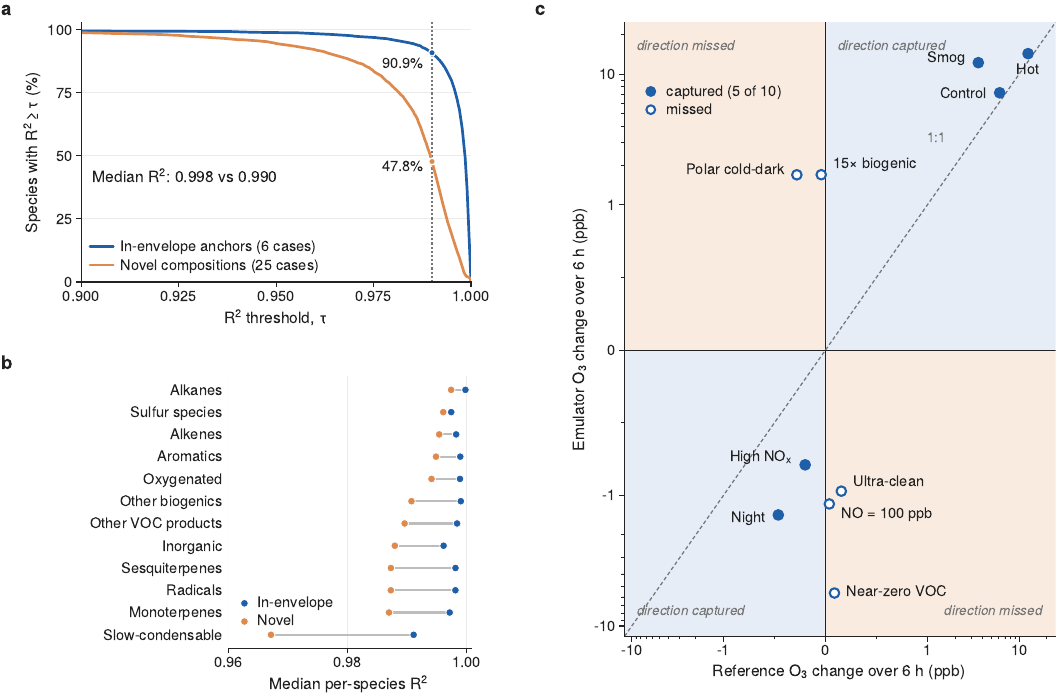}
\caption{\textbf{Skill transfers to novel compositions but not to the most extreme forcing.} \textbf{a}, Share of species whose six-hour per-species $R^2$ exceeds a threshold, for the in-envelope anchors and 25 compositions far from every anchor (physical controls inside the training envelope); the curves separate only in the high-skill tail. \textbf{b}, Median per-species $R^2$ by chemical family for both sets. \textbf{c}, Direction diagram for the ten extreme scenarios, each a realistic base state with one driver pushed to an extreme: six-hour O$_3$ change of 6h-NoStiff against that of the ARCA reference (symmetric-log axes, linear within $\pm1$\,ppb). Points in the shaded same-sign quadrants capture the direction of the reference response (filled, 5 of 10); open circles mark missed directions; dashed, 1:1 line. Time series are shown in Supplementary Fig.~\ref{sfig:extreme}.}\label{fig:generalization}
\end{figure*}

High skill near the training design does not show transfer beyond it. We first built 25 initial compositions far from every anchor while keeping the physical controls inside the training envelope (Methods). Their median standardized distance from the nearest fine-tuning simulation is 1.18, against 0.31 for the evaluation anchors. On these novel compositions the median per-species $R^2$ falls modestly, from 0.998 to 0.990, 98.8\% of species still reach $R^2\geq0.9$, and the median relative error rises from 1.9\% to 3.3\% (Fig.~\ref{fig:generalization}a,b). The clearest change is in the high-skill tail, where the share of species with $R^2\geq0.99$ drops from 90.9\% to 47.8\%. One of the 25 cases diverges.

We then posed ten targeted extreme scenarios, each pushing one driver of a realistic base state to an interpretable extreme (Methods). A physically consistent emulator should reproduce the direction of the ozone response, and FNO-MoE does so in five scenarios (Fig.~\ref{fig:generalization}c; trajectories in \SI{Fig.~\ref{sfig:extreme}}): ozone builds up under base and hot conditions, faster when hotter; decays at night; is titrated under high NO$_x$ (50\,ppb); and rises strongly in the photochemical-smog corner. In the five most aggressive states, the polar cold-dark case at 245\,K, NO of 100\,ppb, near-zero VOC, fifteen-fold biogenic loading and an ultra-clean background, the predicted ozone tendency has the wrong sign. In several of these the true change is small, and the polar cold-dark case is the only clear qualitative miss. In four of the ten runs at least one minor species crosses the divergence criterion, whereas ozone and the other principal diagnostics remain bounded; the median per-species $R^2$ on this set is 0.965. The emulator thus captures canonical responses to strong forcing but does not reliably extrapolate far beyond its training envelope.

\subsection*{Time-varying forcing is the main remaining gap}

Every test so far holds the physical forcing constant within a simulation, whereas in a coupled grid cell it changes between calls. As a first step towards that setting, we drove the six in-envelope compositions with a six-hour morning ramp in short-wave radiation and temperature, which over the rollout rise from about 120 to 850\,W\,m$^{-2}$ and from 288 to 299\,K, supplying the current values to each 30-min block (Methods).

\begin{figure*}[!t]
\centering
\includegraphics[width=\textwidth]{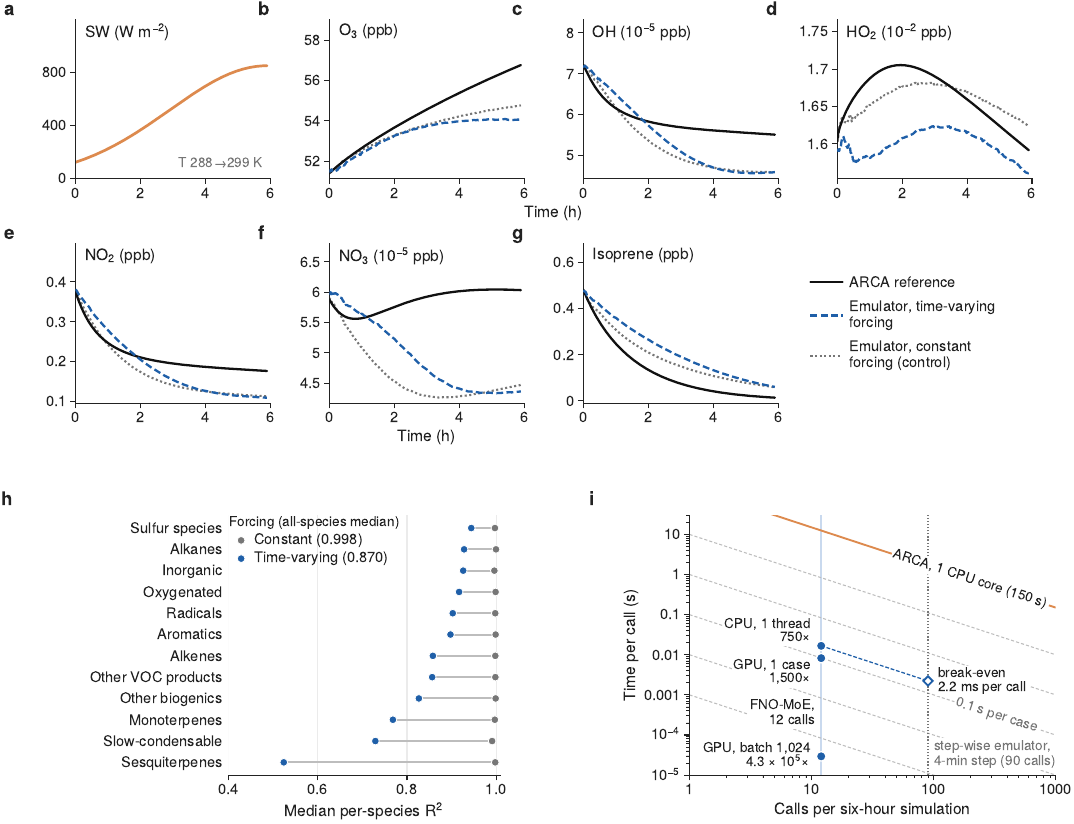}
\caption{\textbf{The emulator does not yet track time-varying forcing, but one call per 30 minutes makes it fast.} \textbf{a}, Short-wave radiation over the six-hour morning ramp (temperature rises in parallel from 288 to 299\,K). \textbf{b}--\textbf{g}, Key species for one representative in-envelope composition: reference (solid), emulator supplied with the time-varying forcing for each block (dashed) and a constant-forcing control (dotted). \textbf{h}, Median per-species $R^2$ by chemical family under constant and time-varying forcing (all-species medians 0.998 and 0.870). \textbf{i}, Cost of a six-hour simulation as time per call against calls per simulation; grey diagonals are constant total time per six-hour case (labelled at 0.1\,s and at the 150-s ARCA integration). FNO-MoE needs 12 calls: 0.20\,s on one single-threaded CPU core (750 times faster than the 150-s single-core ARCA integration), 0.099\,s on one GH200 GPU (1,500 times) and 0.35\,ms per case amortized over a batch of 1,024 ($4.3\times10^{5}$ times). A step-wise emulator with a 4-min step needs 90 calls and would match FNO-MoE on the CPU only if each call took less than 2.2\,ms. Data transfer and coupling overhead are excluded.}\label{fig:forcing}
\end{figure*}

Skill drops markedly (Fig.~\ref{fig:forcing}a--h). The median per-species $R^2$ falls from 0.998 to 0.870, and only 35\% of species stay above 0.9. The loss is largest for reactive biogenic species and their condensable products, whose sinks track the rising oxidant field (monoterpenes 0.77, sesquiterpenes 0.52, slow-condensable products 0.73), and smallest for inorganic and radical families (0.90--0.93). As light increases, the emulator under-predicts the rise in ozone and the mid-morning HO$_2$ peak and the decay of isoprene and monoterpenes.

Supplying the time-varying forcing barely helps. Relative to a control that receives constant forcing, the time-varying rollout is more accurate for four of ten diagnostic species (OH, HO$_2$, NO$_3$, SO$_2$) and less accurate for six, and the two emulator curves nearly coincide. A model trained only on constant-forcing trajectories has therefore not learned to respond to changing controls. One plausible reason is that, under constant forcing, the chemical state already carries the signature of the light and temperature regime, for example in OH and HO$_2$, so the network can predict from the state alone. The correct night-time response in the extreme-scenario test is consistent with this hypothesis, which we have not tested directly. Either way, training on time-varying trajectories is a prerequisite for coupled deployment.

\subsection*{Block prediction makes the emulator fast}

Because one call returns 30 minute-resolved states, six hours take twelve calls; at the 4-min chemistry step of a recent embedded step-wise emulator \cite{liu2025film}, they would take 90. A six-hour reference integration takes about 150\,s on one CPU core. The emulator needs 200\,ms on one single-threaded CPU core, about 750 times faster, and 99\,ms for a single case on one NVIDIA GH200 GPU, about 1,500 times faster (Fig.~\ref{fig:forcing}i). Batching amortizes the fixed GPU overhead: at 1,024 cases, the amortized cost is 0.35\,ms per case, about 2,800 six-hour cases per second or $4.3\times10^{5}$ times the throughput of the single-core reference (\SI{Fig.~\ref{sfig:timing}}). This is throughput, not the latency of one simulation, and not the speedup of a coupled model, which also depends on data movement, other processes and scheduling.

Fewer calls reduce cost only if each call is not correspondingly more expensive. One FNO-MoE call takes about 16\,ms on one CPU core and returns 30 minutes of all 4,691 species, about 0.5\,ms per simulated minute. A step-wise emulator of the same state with a 4-min step needs 7.5 calls per 30 minutes and would match this cost only if each call took less than about 2\,ms. We did not benchmark such an emulator at this state size, so this comparison bounds, rather than measures, the advantage of block prediction.

\section*{Discussion}

We have shown that thousands of near-explicit MCM species can be emulated by a single neural operator that meets, in the box-model setting, the three requirements of an Earth-system chemistry routine. It maps the full retained state and the physical conditions to the next 30 minutes from any point along a trajectory, advances six hours in twelve calls and, after six-hour fine-tuning, remained stable in all 28 tested six-hour rollouts (14 distinct cases for each of two variants). The same network ranked the simulations used to train it. At a fixed budget, this lowered one-step error relative to static and random sampling, on 76\% and 95\% of the evaluation anchors, respectively. The deployed model is accurate across the full species set (median per-species $R^2=0.998$) and on novel compositions ($R^2=0.990$), and it reproduces canonical responses to strong forcing, but it does not reliably extrapolate to the most extreme states or track time-varying forcing ($R^2=0.870$).

The results suggest that one-step accuracy and rollout reliability depend on different properties of the training data. The choice of simulations determines how well the emulator predicts from the states that the design anticipates; trajectory length determines whether it has seen the states that it reaches after hours of self-fed prediction. This is consistent with earlier evidence that training exposure controls accumulated error in online chemical solvers \cite{kelp2022}, and it suggests extending active sampling to score candidates on the states visited during a rollout rather than on the first block alone.

Direct prediction and SIMADy \cite{simady} make complementary trade-offs (\SI{Table~\ref{stab:simady}}). SIMADy evolves six latent species under constrained mass-action dynamics, is bounded by construction and yields an interpretable reduced network. Our emulator predicts every retained species explicitly, and its stability is shown empirically over the tested horizon. A SIMADy-inspired linear autoencoder trained on our data reconstructs the state with about 8.3\% error at every latent dimension from two to ten, about 20 times the 30-min prediction error of 6h-NoStiff in the same units (\SI{Note~\ref{snote:ae}}), which illustrates what a low-dimensional linear code loses for a full-state target. Conversely, the finite-horizon sensitivities of our operator do not concentrate on directly linked reactant--product pairs (\SI{Note~\ref{snote:sens}}), so it is best viewed as a state predictor rather than a reaction model. A constrained reduced model that also predicts a chemically important subset explicitly could combine the strengths of both.

Our study has three groups of limitations, each with a clear remedy. First, the operating range. The emulator is trained, and mostly evaluated, under constant forcing, and its skill drops when the forcing changes within a run. It does not reliably extrapolate far beyond the training envelope, and the envelope-edge test probes only the edges of the training ranges, because the intended extrapolation distances were clipped during data generation. Rollouts also cover six hours, whereas a coupled model integrates for days. Training on time-varying trajectories, regenerating the edge set with widened bounds and multi-day rollouts are therefore the most important next steps.

Second, the strictness of the evaluation. The evaluation anchors are the design anchors, so most tests measure interpolation within the seven-regime design, which the novel-composition test only partly addresses. The regimes are built from domain knowledge rather than fitted to measurements, and every result comes from a single training run. Holding out whole regimes, drawing test conditions from reanalysis or field campaigns, enlarging the six-case in-envelope set and training independent seeds would test generalization and quantify the variability of each gain.

Third, untested aspects of the method. Active sampling was run for one round and compared only at a fixed budget; iterating it, comparing the built-in committee with an independent ensemble \cite{deepensembles} and measuring the number of simulations needed to reach a given accuracy remain to be done. A compute-matched one-hour control and an unrolled training objective would separate the effects of six-hour fine-tuning. Elemental totals are matched only through a soft penalty, for which hard constraints offer a complementary route \cite{beucler2021}. Finally, our baselines do not include a complete SIMADy implementation, a parameter-matched single-expert operator or a step-wise emulator of the same state.

The decisive next test is a coupled experiment that quantifies biases in ozone, HO$_x$, NO$_x$, their reservoirs and condensable oxidation products, together with elemental budgets and end-to-end runtime. A SOSAA column-model experiment with a numerical-solver control would then evaluate the consequences for particle growth and cloud condensation nuclei \cite{sosaa}. Whole-model acceleration will depend on the fraction $f$ of runtime spent in chemistry: for a chemistry speedup $s$ it cannot exceed $1/[(1-f)+f/s]$. With $f=0.95$, the upper end of the range cited above, a model therefore runs at most 20 times faster however fast the chemistry becomes, and $f$ should be reported with kernel timings.

If these tests succeed, near-explicit chemistry could move from box and column models into the settings where its detail matters most: tracing the oxidation pathways that produce secondary organic aerosol and new particles, attributing ozone and OH changes to individual precursors, and running the large ensembles that air-quality and chemistry--climate assessments require. Our results establish the first step in the box-model setting. A neural operator trained on simulations it selected itself predicts thousands of species accurately and stably over repeated calls; making it equally reliable under changing meteorology is what now stands between near-explicit chemistry and the grid cell.

\section*{Methods}

\subsection*{Reference model and training data}

ARCA-box v1.4b serves as the numerical reference; ref.~\cite{arca} describes version 1.2. All experiments except the time-varying test use constant physical controls. Concentrations are saved every 60\,s, giving 61 states for one-hour and 361 states for six-hour simulations. Pressure is fixed at 1013\,hPa and ion production at 3 ion cm$^{-3}$\,s$^{-1}$. The sampled controls are temperature (260--310\,K), relative humidity (0.20--0.99), short-wave radiation (0--900\,W\,m$^{-2}$) and condensation sink ($10^{-4}$--$10^{-2}$\,s$^{-1}$). Each simulation is seeded by a 150-dimensional vector $\theta$ of the four controls and 146 precursor initial concentrations. The vector only initializes the simulation; the emulator receives the complete retained state at the start of each block, and training windows are cut from anywhere along the trajectories. \SI{Table~\ref{stab:datasets}} summarizes the datasets and their roles.

\subsection*{State representation}

Concentrations are converted from molecule\,cm$^{-3}$ to dry-air mixing ratio in ppb with a fixed reference number density (285\,K, 1\,atm), so the conversion factor is constant across simulations, and then transformed as
\begin{equation}
x=\frac{\log_{10}(\mathrm{ppb})+4}{4},
\label{eq:norm}
\end{equation}
so that $\mathrm{ppb}=10^{4x-4}$. Concentrations are floored at $10^{-14}$\,ppb before the logarithm, so non-positive values map to $x=-2.5$. The transform is affine, does not constrain $x$ to $[0,1]$, and outputs are not clipped during training or rollout.

The configured mechanism has 6,068 species (\SI{Fig.~\ref{sfig:network}}). A low-concentration filter removes species whose concentration stays below $10^{-8}$\,ppb in all scanned simulations. Because the filter is applied to each training dataset, the retained count differs slightly: 4,691 for the active-sampling data used by the deployed models, and 4,725 and 4,713 for the random and static data. All-species metrics in the sampling comparison are computed over each model's own retained species; the sets differ by less than 1\% and share 4,673 species, and the key-species metrics use the identical 204 species. Each example consists of an initial retained state $\mathbf{x}_0$, the four controls $\mathbf{m}$ and the 30 subsequent minute-resolved states $\mathbf{X}_{1:H}$.

\subsection*{Fourier mixture-of-experts operator}

The emulator learns the block map
\begin{equation}
\widehat{\mathbf{X}}_{1:H}=\mathcal{G}_{\phi}(\mathbf{x}_0,\mathbf{m}), \qquad H=30\ \mathrm{min},
\end{equation}
in one forward pass, rather than advancing each time step within the network as in formulations that parameterize the state derivative \cite{neural_ode}. Longer integrations feed the final predicted state of one block into the next call.

The core is a one-dimensional Fourier neural operator acting along the output time coordinate \cite{fno,deeponet}. The initial state is broadcast over the horizon and a normalized time coordinate on $[-1,1]$ is appended. Four independent experts each lift this input to 128 channels, apply four spectral blocks and project back to the retained species; they retain 4, 8, 12 and 16 Fourier modes. The four controls condition every spectral block through FiLM \cite{film}. Expert $e$ predicts an increment $\Delta\mathbf{X}^{(e)}$, and a learned static species-wise softmax gate combines the experts \cite{moe}:
\begin{equation}
\begin{gathered}
\widehat{X}_{t,s}=x_{0,s}+\sum_{e=1}^{E}\pi_{s,e}\Delta X_{t,s}^{(e)},\\
\pi_{s,e}=\frac{\exp(g_{s,e})}{\sum_{j=1}^{E}\exp(g_{s,j})}.
\end{gathered}
\end{equation}
The gate is constant across samples and time but varies by species. The model has 7.8 million parameters. All four experts are used by the trained gate, ruling out expert collapse (\SI{Note~\ref{snote:gate}}).

\subsection*{Atmospheric-regime anchors}

Independent sampling over the full 150-dimensional box can combine conditions that do not co-occur in the atmosphere. We therefore construct seven regime archetypes (\SI{Tables~\ref{stab:regimes} and~\ref{stab:ranges}}; \SI{Fig.~\ref{sfig:anchors}}). Anchors are allocated to regimes in fixed quotas proportional to the regime weights (largest-remainder rounding), and Latin-hypercube samples are drawn within each regime's ranges. A regime narrows or shifts selected ranges but never exceeds the global bounds. The 2,500 anchors were generated in two batches: 500 from an earlier four-regime design (150, 150, 100 and 100 anchors) and 2,000 from the seven-regime design, allocated as weight $\times$ 2,000. The same anchors serve for acquisition and for the evaluation in Table~\ref{tab:sampling}. The regimes encode known co-variations: traffic and combustion elevate NO$_x$, CO, aromatics, SO$_2$ and condensation sink; warm, sunlit biogenic conditions elevate isoprene and monoterpenes at low condensation sink; marine conditions combine high humidity with dimethyl sulfide; and polar night has negligible actinic flux. They are constructed from atmospheric-chemistry knowledge rather than estimated from measurements and should be read as a documented design prior.

\subsection*{Sampling strategies and single-round active sampling}

\textbf{Initial condition vectors.} Sampling operates in a normalized space in which each entry of $\theta=(T,\mathrm{RH},\mathrm{SW},\mathrm{CS},c_1,\dots,c_{146})$ is mapped to $[0,1]$ within its global bounds, linearly for temperature, relative humidity and short-wave radiation and logarithmically for condensation sink and concentrations. A candidate around anchor $\mathbf{a}$ lies in the box $[\mathbf{a}-r,\mathbf{a}+r]$, clipped to $[0,1]^{150}$.

\textbf{Matched strategies.} Each strategy uses the same architecture and 10,000 one-hour simulations. Whole-space random sampling draws vectors uniformly within the global bounds. Static local sampling draws Sobol vectors within $r=0.10$ of the anchors and simulates all of them. Active local sampling uses the same anchors, radius and generator but simulates only the candidates selected by the emulator.

\textbf{Warm-up and candidates.} Active sampling starts from 2,500 warm-up simulations, one Sobol draw within each anchor's neighbourhood. The emulator is trained on them and frozen. The anchors are then visited once each in a fixed random permutation (seed 20260803); for anchor $\mathbf{a}_j$, an independent scrambled Sobol draw seeded by the anchor index gives $N=16$ candidates within $r=0.10$. The generator does not use the emulator.

\textbf{Input reconstruction and scoring.} A candidate specifies only precursors and controls, so the emulator input is reconstructed as the state from which the simulation would start: every retained species at the floor and the 146 precursors overwritten with the candidate's values (Eq.~\eqref{eq:norm}), giving $\mathbf{x}_0(\theta)$ and $\mathbf{m}(\theta)$. The acquisition score is the spread of the pre-gate expert increments for these inputs, averaged over the block,
\begin{equation}
u(\theta)=\frac{1}{HS}\sum_{t=1}^{H}\sum_{s=1}^{S}\operatorname*{std}_{e=1,\dots,E}\big[\Delta X^{(e)}_{t,s}\big].
\end{equation}
It uses the experts before the gate, whose weighted average would hide their disagreement, and requires forward passes only. It is necessarily a proxy: before a simulation is run, its initial state is the only state available.

\textbf{Selection, simulation and retraining.} For each anchor the $k=3$ candidates with the largest $u$ are simulated by ARCA and the other 13 discarded, so one sweep scores 40,000 and simulates 7,500 candidates, and warm-up plus sweep exactly fill the 10,000-simulation budget. A fixed quota per anchor, rather than the most uncertain candidates overall, guarantees that every regime receives new simulations. The emulator is then retrained once on all 10,000 simulations, giving 1h-Stiff, from which 1h-NoStiff and the six-hour variants derive.

\textbf{Single-chain alternative.} A preliminary variant explored the space with one random-walk Metropolis chain favouring high disagreement (unit temperature; initial step 0.08 in normalized units, adapted during burn-in). It lingered in a few uncertain neighbourhoods, visiting about 30 of the 2,500 anchors per round, and was discarded (\SI{Table~\ref{stab:acq}}); this comparison does not isolate coverage from the other differences between the algorithms.

\subsection*{Training}

Training minimizes a weighted squared error in normalized-log space with a bias penalty and an elemental term,
\begin{equation}
\begin{aligned}
\mathcal{L}&=\frac{1}{BHS}\sum_{b,t,s} w_s\,(\widehat{X}_{t,s}-X_{t,s})^2\\
&\quad+\lambda_{\mathrm{bias}}\,\mathcal{B}+\lambda_{\mathrm{mass}}\,\mathcal{L}_{\mathrm{mass}},
\end{aligned}
\end{equation}
where $w_s$ equals one except for up-weighted slow-condensable species and selected groups, $\mathcal{L}_{\mathrm{mass}}$ is the elemental term below, $\lambda_{\mathrm{bias}}=0.1$ and $\lambda_{\mathrm{mass}}=0.05$. The bias penalty acts on the up-weighted species,
\begin{equation}
\mathcal{B}=\frac{\sum_s (w_s-1)\,\overline{(\widehat{X}_s-X_s)}^{\,2}}{\max\big(\sum_s\mathbb{1}[w_s>1],\,1\big)},
\end{equation}
where $\overline{\cdot}$ is the mean over batch and horizon. 1h-NoStiff is trained without windows that start in the first five minutes.

\textbf{Six-hour fine-tuning.} 6h-NoStiff and 6h-Stiff continue from 1h-NoStiff and 1h-Stiff on 30-min windows drawn with a one-minute stride from 2,500 six-hour trajectories, using AdamW \cite{adamw} with cosine learning-rate decay. A six-hour simulation yields 331 complete windows against 31 for a one-hour simulation. Checkpoints are selected on a validation split separate from the rollout cases. This is supervised training on windows from longer trajectories, not an unrolled objective, so the gains characterize the complete protocol.

\subsection*{Elemental drift}

With $A_{s,q}$ the number of atoms of element $q$ in species $s$ and $a=4$, $b=-4$,
\begin{equation}
\begin{gathered}
\widehat{M}_{t,q}=\sum_s A_{s,q}\,10^{a\widehat{X}_{t,s}+b},\\
M_{t,q}=\sum_s A_{s,q}\,10^{aX_{t,s}+b},
\end{gathered}
\end{equation}
for $q\in\{\mathrm{C,H,O,N,S}\}$. Elemental drift is $\tfrac{1}{5}\sum_q \big|\log_{10}(\widehat{M}_{t,q}/M_{t,q})\big|$, with both totals clamped at $10^{-8}$. Because the totals use atom counts, they diagnose elemental abundance; agreement with the reference differs from conservation in a closed system, and a small aggregate discrepancy can hide compensating errors among species.

\subsection*{Datasets, splits and test independence}

Training and validation windows are split by reference simulation. A split audit confirms that six-hour fine-tuning does not reuse evaluation conditions: standardizing each dimension of $\theta$ by its spread in the fine-tuning set and measuring root-mean-square distance per dimension, none of the 2,500 evaluation anchors has a near-duplicate among the 2,500 fine-tuning points (minimum nearest-neighbour distance 0.063, median 0.308, none below 0.02), and the six in-envelope rollout cases lie at 0.17--0.40. This separation arises because fine-tuning points are perturbations around anchors, whereas evaluation and rollout cases are the anchors themselves.

The envelope-edge cases hold the 146 initial concentrations at the median-anchor base and set the controls to the edges of the training ranges (\SI{Table~\ref{stab:edge}}). They were generated as 29 runs intended to push the controls 0, 0.25, 0.5 and 1 half-range beyond each edge, but the generator clipped every control to the sampling bounds, so the four distances of each direction produced identical simulations. The test therefore comprises eight distinct cases, the centred control point, six single-control edges and the all-control corner, and all reported envelope-edge statistics except the elemental drift are computed over these eight. Genuine extrapolation of the controls is examined only in the extreme-scenario test, whose bounds were widened. The 25 novel-composition cases are the vectors, from a large whole-space pool with controls held inside the envelope, whose root-mean-square distance to the nearest anchor is largest; in standardized units they lie at a median 1.18 (minimum 1.12) from the nearest fine-tuning point, and at a median 1.17 from the nearest anchor, against 1.02 between neighbouring anchors. The six time-varying cases reuse the in-envelope compositions but drive short-wave radiation and temperature with ARCA's built-in Gaussian ramp,
\[
\mathrm{value}(t)=\min+(\max-\min)\exp\!\left[-\frac{(t-\mu)^2}{2\sigma^2}\right],
\]
with minimum 0\,W\,m$^{-2}$ and 285\,K, maximum 850\,W\,m$^{-2}$ and 299\,K, $\mu$ at the sixth hour and fixed solar angle; the rollout receives the current ramp value per block, and the control receives the window-mean value. All rollout sets are withheld from training and checkpoint selection.

\subsection*{Evaluation metrics}

Errors are computed in normalized-log space (lower is better; zero ideal). For a set $\mathcal{W}$ of 30-min windows, with $X_{w,t,s}$ the value of species $s$ at minute $t$ of window $w$,
\begin{equation}
\mathrm{RMSE}_s^2=\frac{1}{|\mathcal{W}|\,H}\sum_{w\in\mathcal{W}}\sum_{t=1}^{H}\big(\widehat{X}_{w,t,s}-X_{w,t,s}\big)^2,
\end{equation}
and the all-species metric is the root mean square of $\mathrm{RMSE}_s$ over species; the key-species metric restricts this to the 204 key species. Initial, post-transient and full-trajectory metrics use windows starting in the first five minutes, at or after minute five, and at any time. Table~\ref{tab:sampling} pools per-window errors over a middle-of-run interval, whereas the 1h-Stiff/1h-NoStiff comparison uses a per-species evaluation with a fixed five-minute skip, so values from the two pipelines are not directly comparable. In rollouts, a case diverges if any normalized-log value exceeds 3 in magnitude or becomes non-finite. Per-species skill pools all rollout steps and cases: normalized-log RMSE, $R^2=1-\sum(\widehat{X}-X)^2/\sum(X-\bar{X})^2$ in normalized-log space, and mean relative error $\lvert\widehat{c}-c\rvert/(\lvert c\rvert+\epsilon)$ in ppb. All values come from single training runs.

\subsection*{Extreme scenarios}

Ten scenarios start from the per-parameter median of the 2,500 anchors and push one driver to an extreme: zero short-wave radiation (night; with 245\,K, polar cold-dark); NO of 50 and 100\,ppb; 325\,K; all organic precursors near zero; fifteen-fold biogenic loading; a hot, high-NO$_x$, aromatic-rich, full-sun corner (smog); and an ultra-clean low-NO$_x$, low-VOC background. ARCA integrates each for six hours under constant controls, with sampling bounds widened so that extreme values are not clipped, and the deployed model is rolled out with the in-envelope protocol.

\subsection*{Reduced-order representation baseline}

The SIMADy-inspired baseline \cite{simady} is a tied-weight linear autoencoder with non-negative encoder $E\in\mathbb{R}^{L\times S}$ that reconstructs a standardized concentration vector $z$ as $E^{\top}Ez$. The loss is the reconstruction mean-square error, with 100-fold weight on O$_3$, OH, HO$_2$, NO, NO$_2$ and NO$_3$, plus an off-diagonal latent-covariance penalty (coefficient 1.0). We train with Adam \cite{adam} (learning rate $3\times10^{-4}$, 400 epochs) for $L\in\{2,4,6,8,10\}$. Non-negativity is enforced by projection after each step, and the emissions-reconstruction term is omitted because the simulations have no emission input.

\subsection*{Timing}

Emulator timings cover a full six-hour integration of twelve 30-min blocks and are compared with a six-hour ARCA integration on one CPU core (median wall-clock time in the generation logs). CPU timings restrict PyTorch to one thread; GPU timings are reported for one case and amortized over batches of up to 1,024. Inputs are float32 and already resident on the device, so data transfer and preprocessing are excluded. Each value is the mean of 20 timed repetitions after three warm-up repetitions, with GPU synchronization, on one node with an NVIDIA GH200 superchip.

\section*{Data availability}

The ARCA input configurations, generated trajectories, anchor definitions, trained checkpoints and evaluation outputs will be deposited in a permanent repository [TODO: DOI and licence]. Source data for Figs.~2--6 are provided with this paper.

\section*{Code availability}

The training, evaluation and figure code will be released under [TODO: licence] at [TODO: repository URL] and archived with a DOI; it is available to editors and reviewers during peer review.

\section*{Acknowledgements}

[TODO: funding and computing resources, e.g. CSC -- IT Center for Science, Finland.]

\section*{Author contributions}

[TODO: CRediT contributions.]

\section*{Competing interests}

[AUTHOR CONFIRMATION REQUIRED: competing-interests statement approved by all authors.]

\section*{Additional information}

Supplementary Information, comprising Supplementary Notes~\ref{snote:ae}--\ref{snote:key}, Supplementary Tables~\ref{stab:regimes}--\ref{stab:simady} and Supplementary Figs~\ref{sfig:anchors}--\ref{sfig:extreme}, follows the references.

\bibliography{main}

\clearpage
\section*{Supplementary Information}

\setcounter{figure}{0}
\setcounter{table}{0}
\renewcommand{\figurename}{Supplementary Fig.}
\renewcommand{\tablename}{Supplementary Table}
\renewcommand{\theHfigure}{S\arabic{figure}}
\renewcommand{\theHtable}{S\arabic{table}}

\snote{a low-dimensional linear code discards much of the retained state}{snote:ae}

To place direct prediction alongside reduced-order approaches, we trained the SIMADy-inspired tied-weight, non-negative linear autoencoder described in Methods on the same ARCA trajectories. This diagnostic measures how well one low-dimensional linear code reconstructs the state; it does not measure the accuracy, stability or runtime of learned latent dynamics. The autoencoder reconstructs all species, and the key species, with approximately 8.3\% error relative to RMS concentration, essentially independent of the latent dimension from two to ten (Supplementary Fig.~\ref{sfig:ae}). Ozone, which is up-weighted during training, is reconstructed to 5--6\% with four to eight latent variables, but less well at the ends of the range (31\% with two and 17\% with ten). In the same units, the 30-min prediction error of 6h-NoStiff is 0.36\% for all species and 0.95\% for ozone, about 20 times lower than the reconstruction error for all species. Reconstruction and prediction are different tasks, so this comparison shows the cost of the tested representation rather than ranking complete methods, and the plateau is not a lower bound for larger, nonlinear or dynamically trained latent models.

\snote{learned sensitivities do not map onto individual reaction links}{snote:sens}

The configured mechanism links 6,068 species through 18,490 reactions (Supplementary Fig.~\ref{sfig:network}). The operator receives no reaction adjacency; the only mechanism information it uses is elemental composition, through the elemental term of the training loss. Its finite-horizon sensitivities, $|\partial\widehat{X}_i/\partial x_{0,j}|$, respond broadly to inorganic and radical species but do not concentrate systematically on directly linked reactant--product pairs (Supplementary Fig.~\ref{sfig:jacobian}). A finite-time flow-map derivative includes indirect pathways and is not the instantaneous chemical Jacobian, so this result neither establishes nor rules out chemical structure in the learned dynamics.

\snote{the expert gate does not collapse}{snote:gate}

A mixture of experts can collapse onto a subset of experts, leaving the rest untrained. All four experts of the deployed model carry near-uniform mean weight (0.247--0.261, against 0.25 for a uniform gate), and each is the highest-weighted expert for a substantial block of species (560--2,582 of 4,691), so none is unused (Supplementary Fig.~\ref{sfig:gate}). The per-species gate is close to uniform (mean normalized entropy 0.99), meaning that the experts contribute jointly rather than hard-routing; the highest-weighted expert nonetheless tilts by chemical family, for example towards the high-mode expert for monoterpenes and inorganics and towards the second expert for radicals. The pre-gate disagreement that active sampling uses as its score remains informative regardless of the near-uniform gate.

\snote{key species}{snote:key}

The 204 key species used for the key-species metrics, the intersection of a domain-curated list of tracked species with the retained species, are listed below in MCM nomenclature.

{\small\noindent ACO2H, ACR, ALLYLOH, APINENE, BCARY, BENZAL, BENZENE, BIACET, BOX2PROL, BPINENE, BUOX2ETOH, BUT1ENE, BUT2CHO, BUT2OL, BUT2OLO, BUTACID, C2H2, C2H4, C2H5CHO, C2H5OH, C2H6, C3H6, C3H7CHO, C3H8, C3ME3CHO, C3ME3CO2H, C4ALDB, C4H6, C4H9CHO, C5H11CHO, C5H8, C6H13CHO, C6H13CO2H, C6H5CH2OH, CATECHOL, CBUT2ENE, CCL2CH2, CDICLETH, CH2CL2, CH2CLCH2CL, CH2CLCHCL2, CH3BR, CH3CCL3, CH3CH2CL, CH3CHO, CH3CL, CH3CO2H, CH3COCH3, CH3O2, CH3OCH3, CH3OCHO, CH3OH, CH3OOH, CH3SO2, CH3SO3, CH4, CHCL2CH3, CHCL2CHCL2, CHCL3, CHEX, CHEX2ENE, CL12PROP, CO, CO24C5, CPENT2ENE, CRESOL, CYHEXOL, CYHEXONE, DIBRET, DIEK, DIET35TOL, DIETETHER, DIME35EB, DM23BU2ENE, DMA, DMC, DMM, DMS, DMSO, DMSO2, DNCRES, DNPHEN, EBENZ, EBENZOL, EGLYOX, EOX2EOL, ETBE, ETHACET, ETHGLY, ETHOX, GLYOX, H2, H2O2, H2SO4, HCHO, HCL, HCOOH, HEX1ENE, HEX2ONE, HEX3ONE, HEXAOH, HNO3, HO2, HOC6H4NO2, HONO, IBUTACID, IBUTOL, IC4H10, IC5H12, IPBENZ, IPEAOH, IPEBOH, IPECOH, IPRCHO, IPROACET, IPROPOL, LIMONENE, M22C4, M23C4, M2HEX, M2PE, M2PEBOH, M3F, M3HEX, M3PE, MACR, MALANHY, MBO, ME2BUT1ENE, ME2BUT2ENE, ME3BUOL, ME3BUT1ENE, MEK, MEPROPENE, METHACET, METHCOACET, METHTOL, MGLYOX, MIBK, MIBKAOH, MIPK, MO2EOL, MPRK, MSA, MTBE, MTBK, MXYL, MXYLAL, N2O5, NBUTACET, NBUTOL, NC10H22, NC11H24, NC12H26, NC4H10, NC5H12, NC6H14, NC7H16, NC8H18, NC9H20, NEOP, NH3, NO, NO2, NO3, NPROACET, NPROPOL, O3, OETHTOL, OH, OXYL, OXYLAL, PAN, PBENZ, PEBOH, PECOH, PENT1ENE, PENTACID, PETHTOL, PHCOME, PHCOOH, PHENOL, PHIC3OH, PR2OHMOX, PROPACID, PROPGLY, PXYL, SBUTACET, SO2, STYRENE, TBUACET, TBUT2ENE, TBUTOL, TCE, TDICLETH, THEX2ENE, TM123B, TM123BCHO, TM124B, TM135B, TOLUENE, TPENT2ENE, TRICLETH, VINCL}

\begin{table*}[!t]
\caption{\textbf{Seven hand-constructed atmospheric regimes define the anchor design.} The weight sets each regime's share of anchors. The anchor counts are fixed quotas from two generation batches, 500 anchors from an earlier four-regime design and 2,000 from the seven-regime design allocated as weight $\times$ 2,000, and therefore differ from weight $\times$ 2,500. The same anchors are used for acquisition and evaluation; regime-specific ranges are listed in Supplementary Table~\ref{stab:ranges}.}\label{stab:regimes}
\centering
\tablestyle
\begin{tabular}{lcc}
\toprule
\textbf{Regime} & Sampling weight & Evaluation anchors \\
\midrule
Clean biogenic summer day & 0.20 & 550 \\
Urban polluted & 0.20 & 550 \\
Cold low-biogenic winter night & 0.15 & 400 \\
Mixed background & 0.15 & 400 \\
Marine remote background & 0.15 & 300 \\
Biomass-burning plume & 0.10 & 200 \\
Polar pristine night & 0.05 & 100 \\
\bottomrule
\end{tabular}
\end{table*}

\begin{table*}[!t]
\caption{\textbf{Regime-specific parameter ranges for the anchor design.} Sampling range within each regime for the four physical controls and nine signature precursors; a dash means the regime inherits the global range. Ranges are hand-constructed from atmospheric-chemistry relationships, not fitted to measurements, and are clamped inside the global bounds. Controls: temperature $T$, short-wave radiation SW, relative humidity RH and condensation sink CS (ARCA inputs TEMPK, SW\_RADIATION, REL\_HUMIDITY and CONDENS\_SINK). Global ranges, used where a regime does not override and for whole-space random sampling: $T$ 260--310\,K; SW 0--900\,W\,m$^{-2}$; RH 0.2--0.99; CS $10^{-4}$--$10^{-2}$\,s$^{-1}$; and, in ppb, NO 0.01--10, NO$_2$ 0.05--10, O$_3$ 10--80, CO 50--500, SO$_2$ 0.02--5, $\alpha$-pinene 0.01--2, isoprene 0.01--5, toluene 0.01--5 and DMS 0.0022--0.22. The full 146-species precursor vector carries further regime overrides, given in the released configuration.}\label{stab:ranges}
\centering
\tablestyle\footnotesize\setlength{\tabcolsep}{3pt}
\begin{tabular}{lccccccc}
\toprule
\multirow{2}{*}{\textbf{Parameter}} & \multicolumn{7}{c}{\textbf{Regime}} \\
\cmidrule(l){2-8}
 & Clean & Urban & Winter & Mixed & Marine & Biomass & Polar \\
\midrule
Design weight & 0.20 & 0.20 & 0.15 & 0.15 & 0.15 & 0.10 & 0.05 \\
\grouprow{8}{Physical controls}
$T$ (K) & 290--310 & -- & 260--280 & -- & -- & 285--305 & 260--270 \\
SW (W\,m$^{-2}$) & 500--900 & -- & 0--100 & -- & -- & 300--900 & 0--5 \\
RH & 0.2--0.55 & -- & 0.6--0.99 & -- & 0.7--0.99 & 0.2--0.5 & 0.6--0.99 \\
CS ($10^{-3}$\,s$^{-1}$) & 0.1--1 & 1--10 & 0.5--10 & -- & 0.1--1 & 3--10 & 0.1--1.5 \\
\grouprow{8}{Precursor concentrations (ppb)}
NO & 0.01--0.5 & 1--10 & 0.3--5 & -- & 0.01--0.3 & 0.3--5 & 0.01--0.2 \\
NO$_2$ & 0.05--0.5 & 1--10 & 0.3--5 & -- & 0.05--0.5 & 0.3--5 & 0.05--0.3 \\
O$_3$ & 40--80 & 10--50 & 10--30 & -- & 10--40 & 20--60 & 15--35 \\
CO & 50--200 & 150--500 & 100--500 & -- & 50--120 & 200--500 & 60--150 \\
SO$_2$ & 0.02--0.3 & 0.5--5 & 0.1--2 & -- & 0.02--0.5 & 0.02--1 & 0.02--1 \\
$\alpha$-Pinene & 0.3--2 & 0.01--0.3 & 0.01--0.05 & -- & 0.01--0.05 & 0.02--0.5 & 0.01--0.03 \\
Isoprene & 0.5--5 & 0.01--0.8 & 0.01--0.1 & -- & 0.01--0.1 & 0.05--1.5 & 0.01--0.05 \\
Toluene & 0.01--0.3 & 0.5--5 & 0.05--2 & -- & 0.01--0.1 & 0.3--3 & 0.01--0.1 \\
DMS & -- & -- & -- & -- & 0.05--0.22 & -- & -- \\
\bottomrule
\end{tabular}
\end{table*}

\begin{table*}[!t]
\caption{\textbf{Reference-trajectory datasets and their roles.}}\label{stab:datasets}
\centering
\tablestyle
\begin{tabular}{lrrl}
\toprule
\textbf{Dataset} & Simulations & Duration & Role \\
\midrule
\grouprow{4}{Training and fine-tuning}
Whole-space random & 10,000 & 1 h & Random-sampler training \\
Static local sampling & 10,000 & 1 h & Static-sampler training \\
Active local sampling & 10,000 & 1 h & 1h-Stiff and 1h-NoStiff training \\
Long-trajectory anchors & 2,500 & 6 h & 6h-NoStiff and 6h-Stiff fine-tuning \\
\grouprow{4}{Evaluation}
Regime anchors & 2,500 & 1 h & One-step evaluation \\
In-envelope rollout set & 6 & 6 h & Autoregressive test \\
Envelope-edge rollout set & 8 (29 runs) & 6 h & Envelope-edge test \\
Novel-composition set & 25 & 6 h & Generalization test \\
Extreme-scenario set & 10 & 6 h & Physical-response test \\
Time-varying set & 6 & 6 h & Time-varying forcing test \\
\bottomrule
\end{tabular}
\end{table*}

\begin{table*}[!t]
\caption{\textbf{Construction of the envelope-edge rollout cases.} The generator was designed to push the physical controls beyond the training envelope by a distance $d$ (half-ranges beyond the edge for the linear controls, decades for the condensation sink), one control at a time plus an all-control corner, with the 146 initial concentrations at the median-anchor base. Every control was, however, clipped to the sampling bounds, so all four distances of a direction produced the same simulation at the edge. The 29 runs therefore contain eight distinct cases: the centred control point (285\,K, RH 0.595, 450\,W\,m$^{-2}$, $5.05\times10^{-3}$\,s$^{-1}$) and the seven edge settings listed below.}\label{stab:edge}
\centering
\tablestyle
\begin{tabular}{llcc}
\toprule
\textbf{Control} & Training envelope & Direction & Simulated value (all $d$) \\
\midrule
Temperature & 260--310\,K & low, high & 260 / 310\,K \\
Relative humidity & 0.2--0.99 & low & 0.2 \\
Short-wave radiation & 0--900\,W\,m$^{-2}$ & high & 900\,W\,m$^{-2}$ \\
Condensation sink & $10^{-4}$--$10^{-2}$\,s$^{-1}$ & low, high & $10^{-4}$ / $10^{-2}$\,s$^{-1}$ \\
Corner (all four) & --- & outward & 310\,K, 0.2, 900\,W\,m$^{-2}$, $10^{-2}$\,s$^{-1}$ \\
\bottomrule
\end{tabular}
\end{table*}

\begin{table*}[!t]
\caption{\textbf{Coverage-guaranteed active sampling outperforms a single-chain Metropolis strategy.} Overall normalized-log RMSE (all species, full trajectory) at a matched 10,000-simulation budget; this metric is shown as vertical lines in Supplementary Fig.~\ref{sfig:errdist}. Only the scalar RMSE was retained for the single-chain strategy, which visited about 30 of the 2,500 anchors per round. Lower is better; best value in bold.}\label{stab:acq}
\centering
\tablestyle
\begin{tabular}{lc}
\toprule
\textbf{Acquisition strategy} & Overall RMSE\lowerbetter \\
\midrule
\rowcolor{tabbase} Whole-space random & 0.0528 \\
\rowcolor{tabbase} Single-chain Markov-chain Monte Carlo & 0.0335 \\
\rowcolor{tabbase} Static local sampling & 0.0308 \\
\rowcolor{tabours} \textbf{Active local sampling (best-of-$N$ disagreement)} & \textbf{0.0271} \\
\bottomrule
\end{tabular}
\end{table*}

\begin{table*}[!t]
\caption{\textbf{Direct prediction and SIMADy make complementary design choices.} Methodological distinctions, not a matched performance ranking.}\label{stab:simady}
\centering
\tablestyle
\begin{tabular}{>{\raggedright\arraybackslash}p{0.20\textwidth}>{\raggedright\arraybackslash}p{0.34\textwidth}>{\raggedright\arraybackslash}p{0.34\textwidth}}
\toprule
 & \textbf{SIMADy} \cite{simady} & \textbf{This work} \\
\midrule
Reference & PartMC--MOSAIC with MCM & ARCA-box with MCM \\
Representation & Six latent species & 4,691 retained species as direct outputs \\
Dynamics & Sparse mass-action system, integrated in time & Fourier mixture of experts, 30 states per call \\
Training data & Latin-hypercube scenarios & Emulator-selected simulations \\
Stability & Bounded by construction & Empirical six-hour rollout tests \\
Evaluation & Time-varying multi-day forcing & Constant physical controls; one time-varying test \\
\bottomrule
\end{tabular}
\end{table*}

\begin{figure*}[p]
\centering
\includegraphics[width=\textwidth,height=0.57\textheight,keepaspectratio]{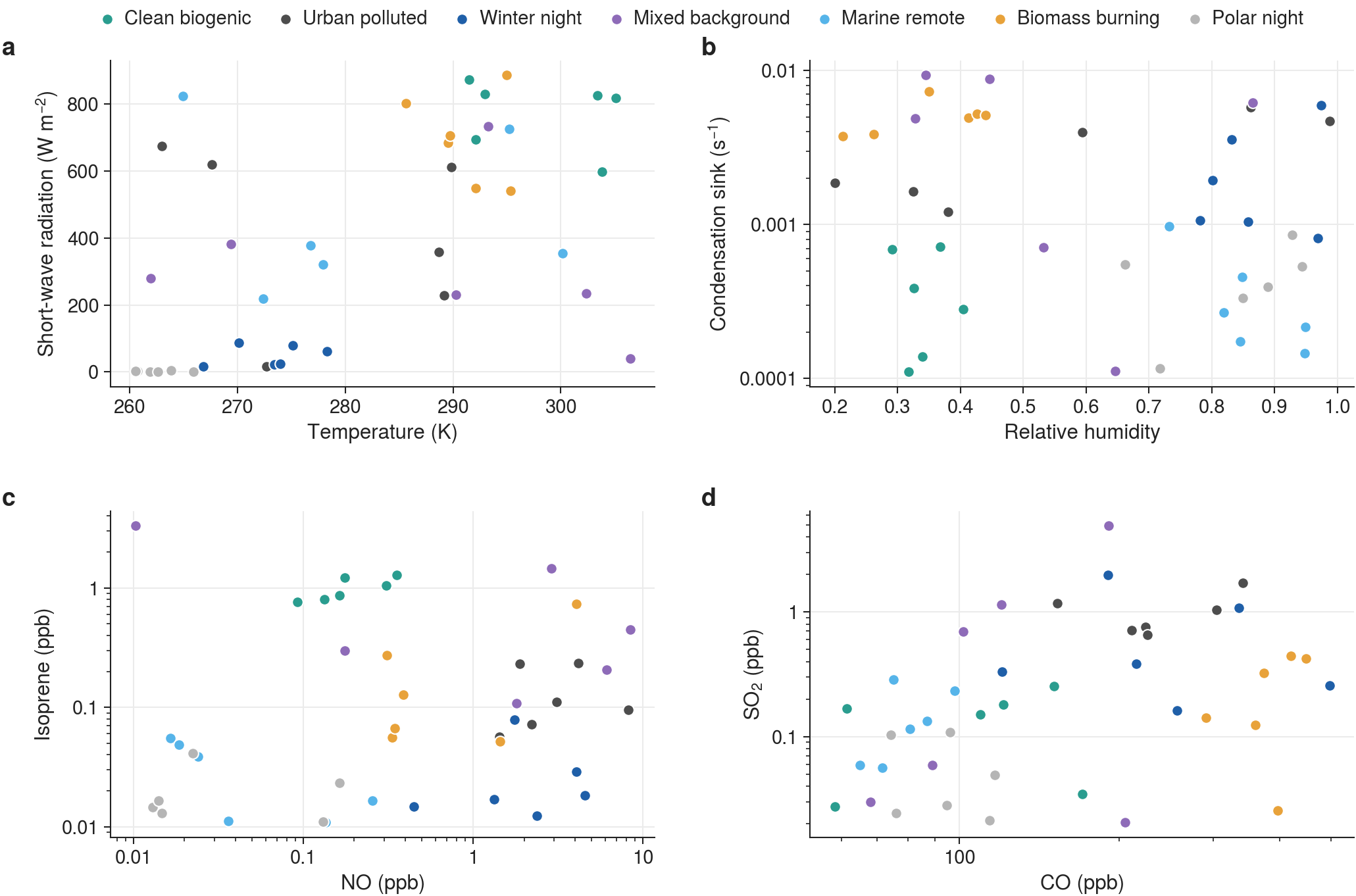}
\caption{\textbf{Regime anchors combine physical drivers and precursor mixtures that co-vary in the atmosphere.} Six reproducible draws for each of the seven hand-constructed regimes, showing the joint ranges of physical drivers (a,b) and chemically informative initial precursors (c,d). Variables without a regime-specific range keep the global bounds. The panels document the sampling design, not a measurement-derived distribution.}\label{sfig:anchors}
\end{figure*}

\begin{figure*}[p]
\centering
\includegraphics[width=\textwidth,height=0.57\textheight,keepaspectratio]{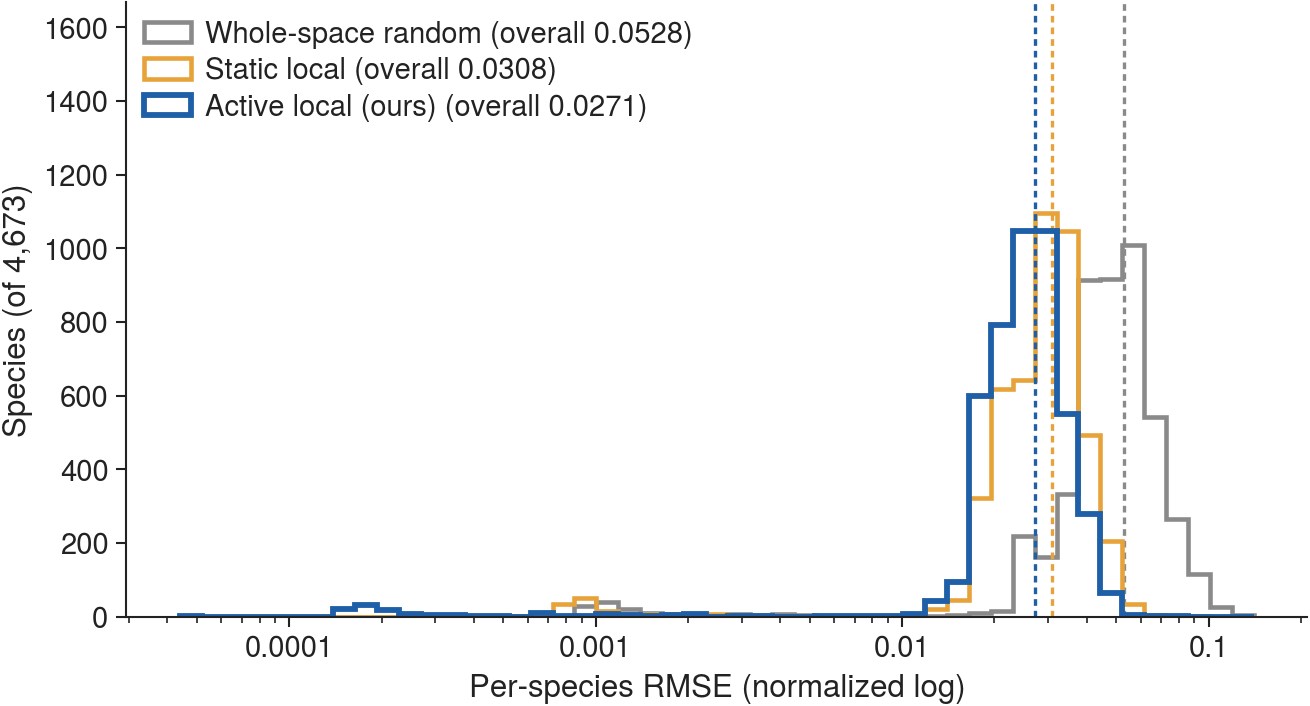}
\caption{\textbf{Active sampling shifts the per-species error distribution towards lower error.} Histograms of per-species normalized-log RMSE for the three 10,000-simulation strategies; vertical lines mark each strategy's overall RMSE (Supplementary Table~\ref{stab:acq}). Paired species-level differences would be needed to identify individual species that regress.}\label{sfig:errdist}
\end{figure*}

\begin{figure*}[p]
\centering
\includegraphics[width=\textwidth,height=0.57\textheight,keepaspectratio]{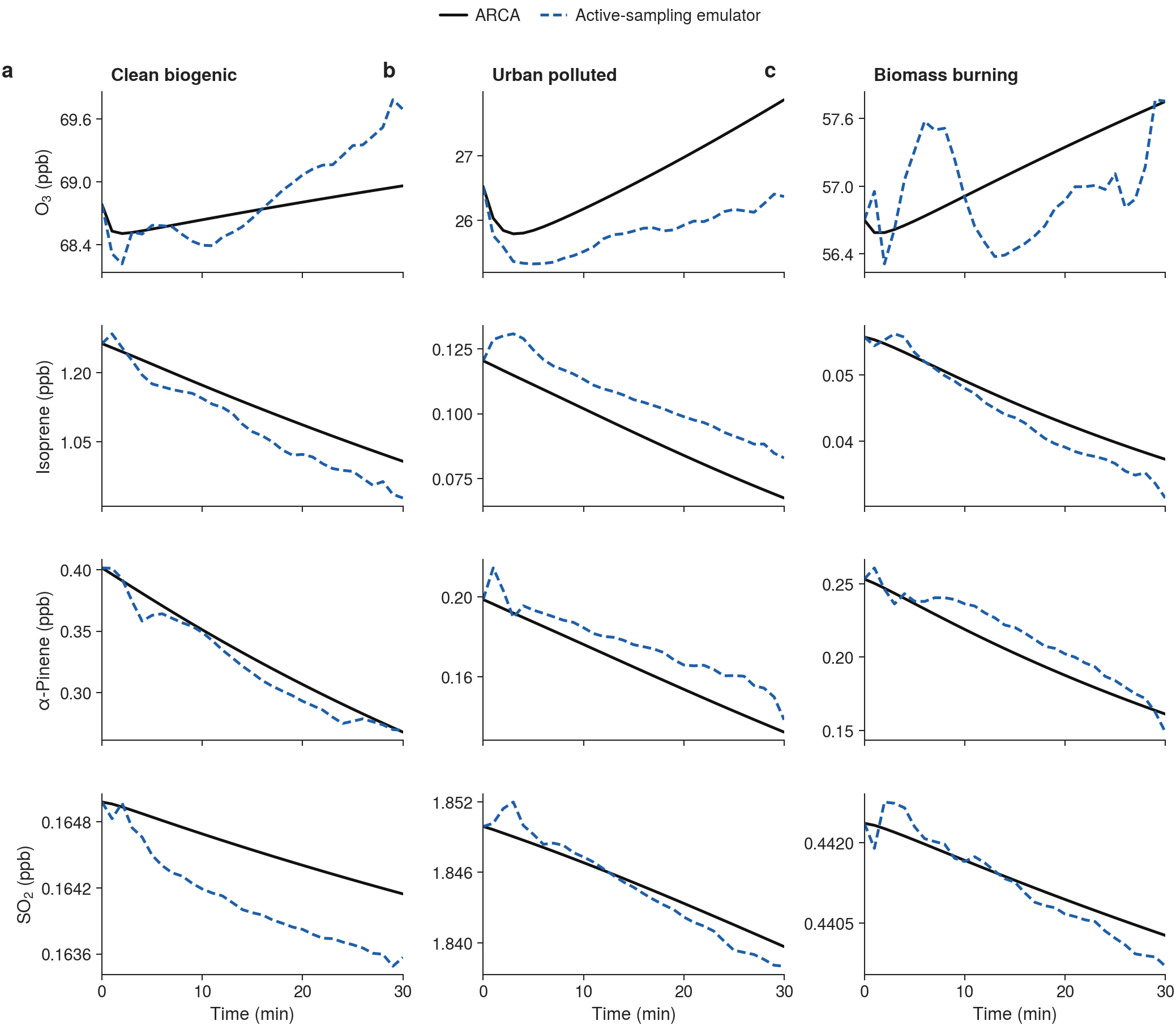}
\caption{\textbf{The emulator reproduces the principal precursor decay at evaluation anchors.} Minute-resolved ARCA solutions (solid black) and active-sampling emulator predictions (dashed blue) from the same initial retained state. Columns: clean-biogenic, urban-polluted and biomass-burning anchors; rows: O$_3$, isoprene, $\alpha$-pinene and SO$_2$ (ppb). Deviations include the curvature of O$_3$. Illustrative traces, not an additional accuracy metric.}\label{sfig:odefits}
\end{figure*}

\begin{figure*}[p]
\centering
\includegraphics[width=\textwidth,height=0.57\textheight,keepaspectratio]{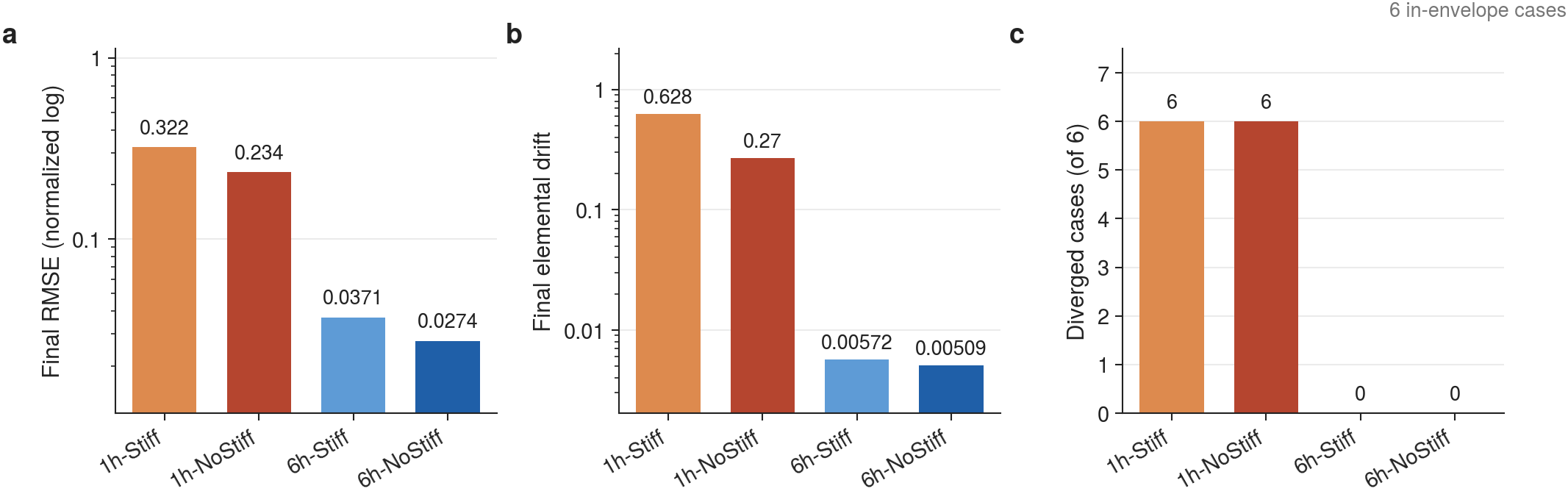}
\caption{\textbf{Six-hour fine-tuning prevents divergence in the in-envelope rollouts.} Final normalized-log RMSE (a), elemental drift (b) and number of diverging cases out of six (c) for the four variants. Panels a and b use logarithmic axes.}\label{sfig:interp}
\end{figure*}

\begin{figure*}[p]
\centering
\includegraphics[width=\textwidth,height=0.57\textheight,keepaspectratio]{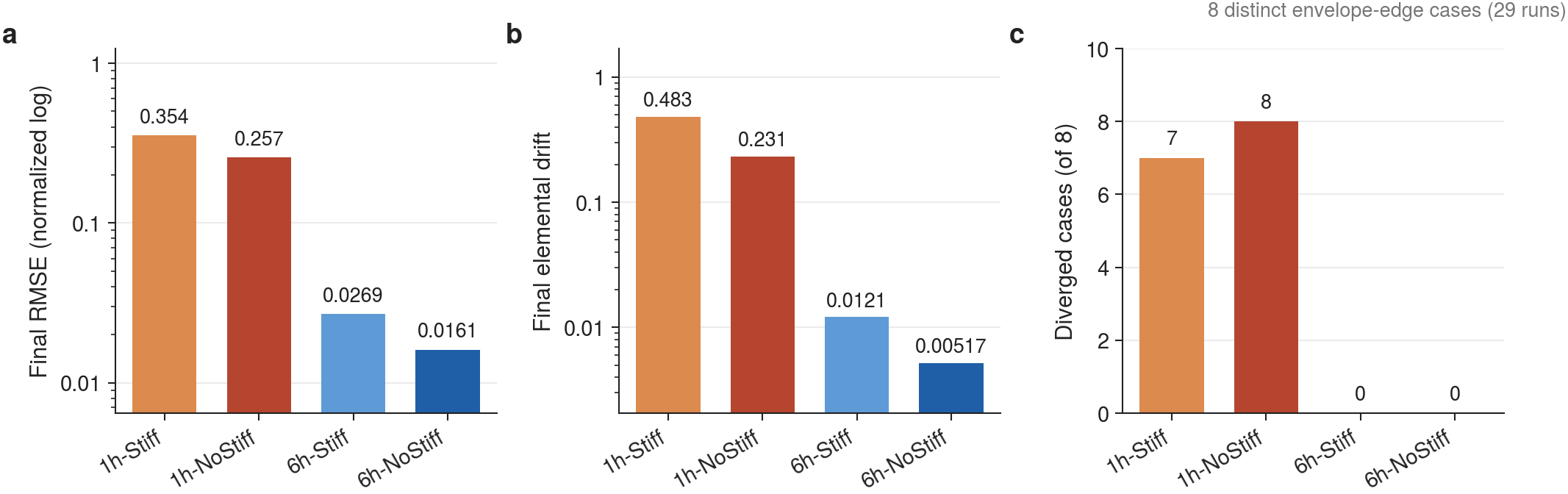}
\caption{\textbf{The stability gain persists at the edges of the training envelope.} As Supplementary Fig.~\ref{sfig:interp}, for the 29 envelope-edge runs, which comprise eight distinct cases because the intended extrapolation distances were clipped to the sampling bounds (Supplementary Table~\ref{stab:edge}). RMSE (a) and divergence counts (c) are over the eight distinct cases (7/8, 8/8, 0/8 and 0/8); the elemental drift (b) is averaged over the 29 runs because per-case drift was not stored.}\label{sfig:extrap}
\end{figure*}

\begin{figure*}[p]
\centering
\includegraphics[width=\textwidth,height=0.57\textheight,keepaspectratio]{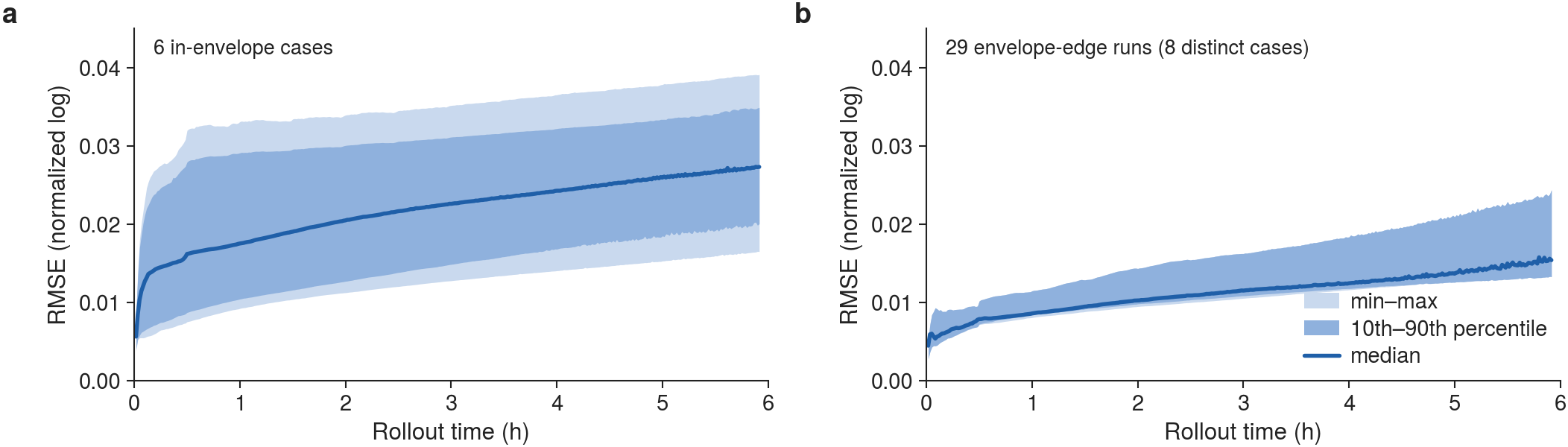}
\caption{\textbf{Rollout error grows gradually over six hours.} Per-case normalized-log RMSE of 6h-NoStiff versus rollout time for six in-envelope cases (a) and the 29 envelope-edge runs (b; eight distinct cases). Solid line, median; shading, minimum--maximum range and empirical 10th--90th percentiles. These ranges describe the tested cases and are not confidence intervals; the lower aggregate error at the envelope edges should not be read as evidence that extrapolation is easier.}\label{sfig:percentiles}
\end{figure*}

\begin{figure*}[p]
\centering
\includegraphics[width=\textwidth,height=0.62\textheight,keepaspectratio]{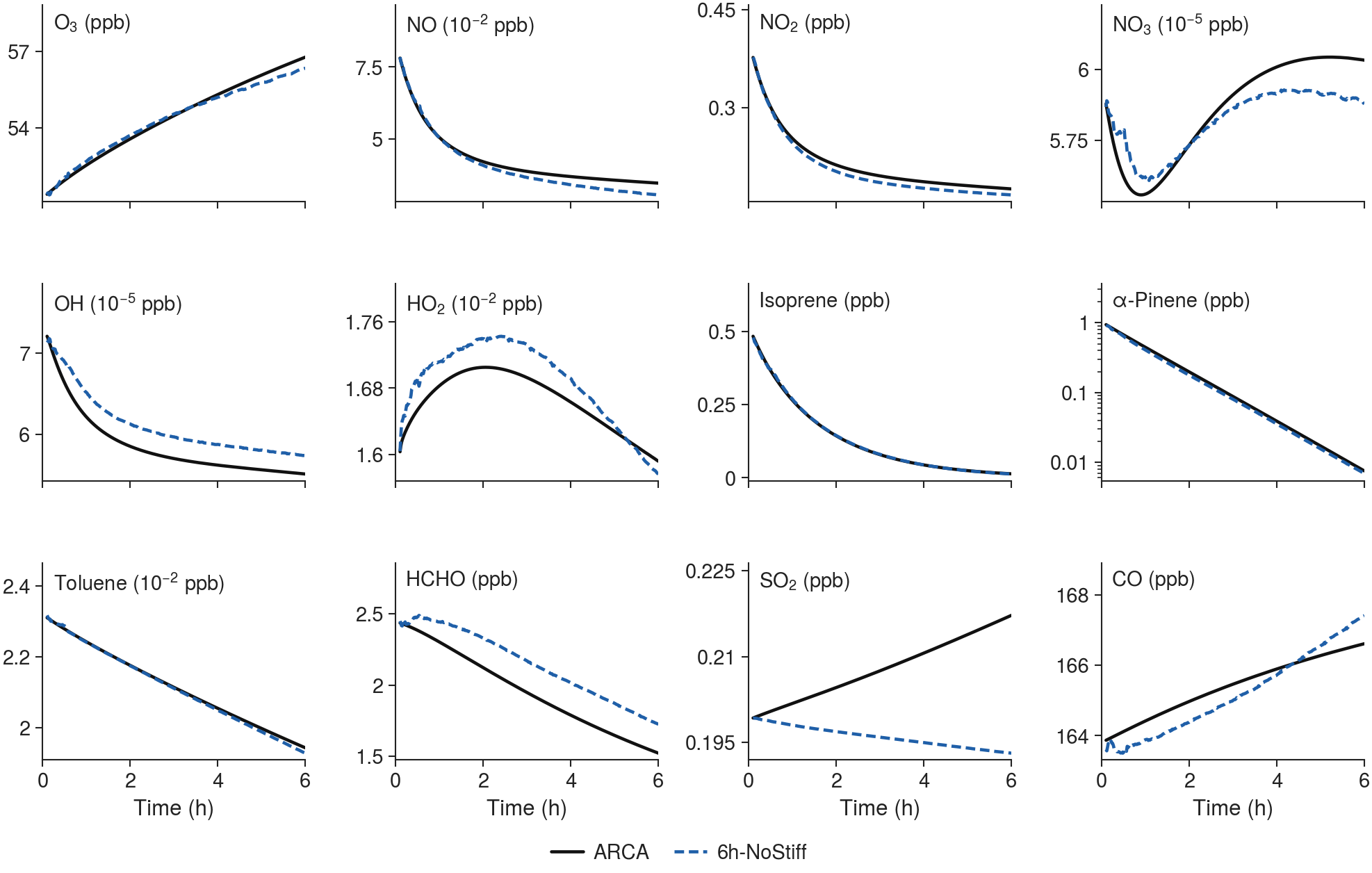}
\caption{\textbf{Six-hour rollout trajectories reproduce chemistry across families.} Predicted (6h-NoStiff, dashed) and reference (ARCA, solid) concentrations over a representative in-envelope six-hour simulation for twelve species spanning inorganic, radical, biogenic, aromatic, oxygenated and sulfur families. Axes are logarithmic where the dynamic range exceeds a factor of 40.}\label{sfig:traj}
\end{figure*}

\begin{figure*}[p]
\centering
\includegraphics[width=\textwidth,height=0.42\textheight,keepaspectratio]{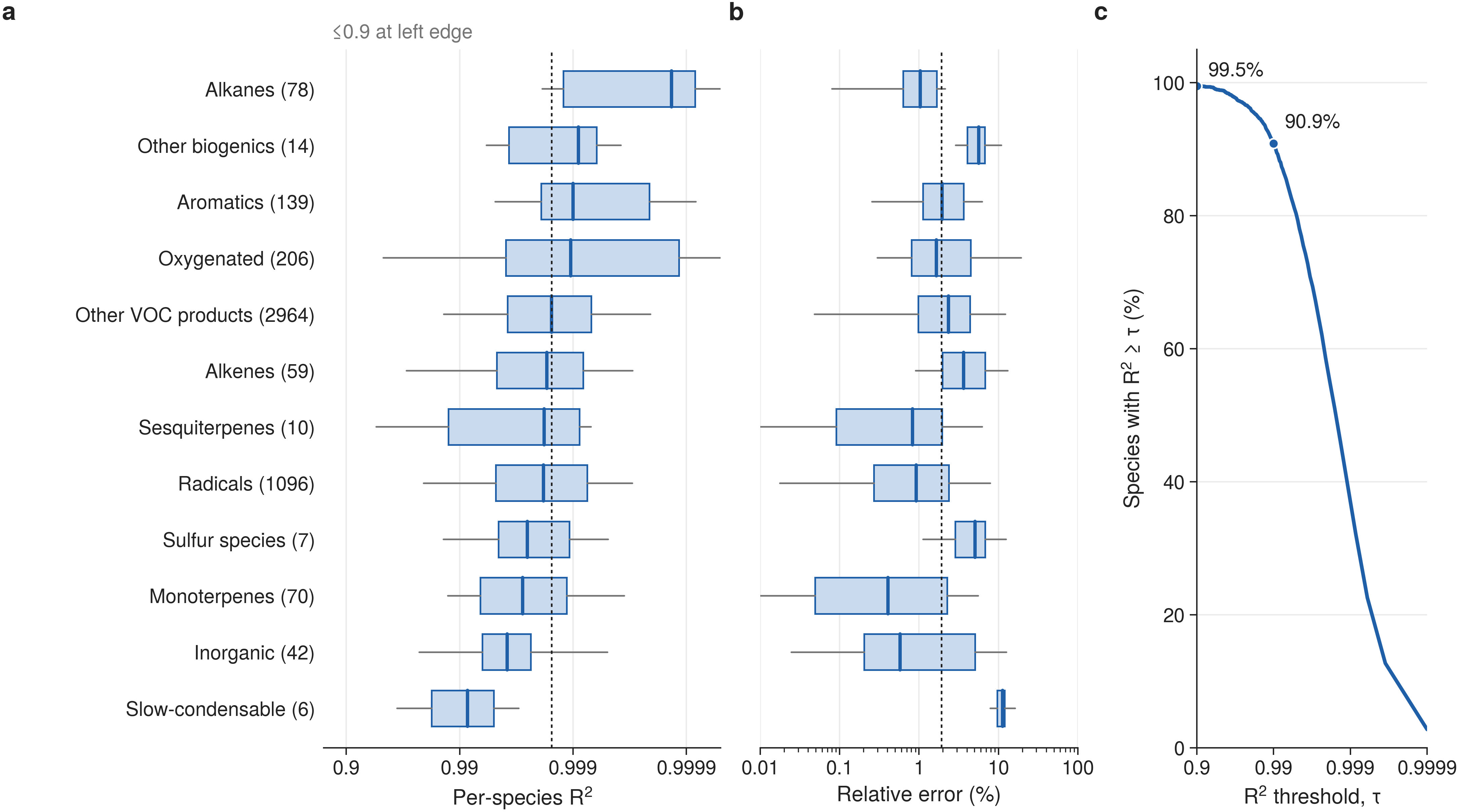}
\caption{\textbf{Skill is high in every chemical family.} \textbf{a}, Per-species $R^2$ (normalized-log space, six-hour rollouts of 6h-NoStiff) by chemical family (species count in brackets); the axis is scaled as $-\log_{10}(1-R^2)$ as in Fig.~\ref{fig:fullstate}b, and values below 0.9 are drawn at the left edge. \textbf{b}, Per-species relative concentration error (\% of reference ppb) by family, logarithmic axis. Boxes, interquartile range; line, median; whiskers, 5th--95th percentiles; dotted lines, all-species medians (0.998 and 1.9\%). \textbf{c}, Percentage of all 4,691 retained species whose $R^2$ exceeds a threshold $\tau$ (99.5\% at 0.9, 90.9\% at 0.99). Relative errors are largest for low-abundance families.}\label{sfig:skill}
\end{figure*}

\begin{figure*}[p]
\centering
\includegraphics[width=\textwidth,height=0.5\textheight,keepaspectratio]{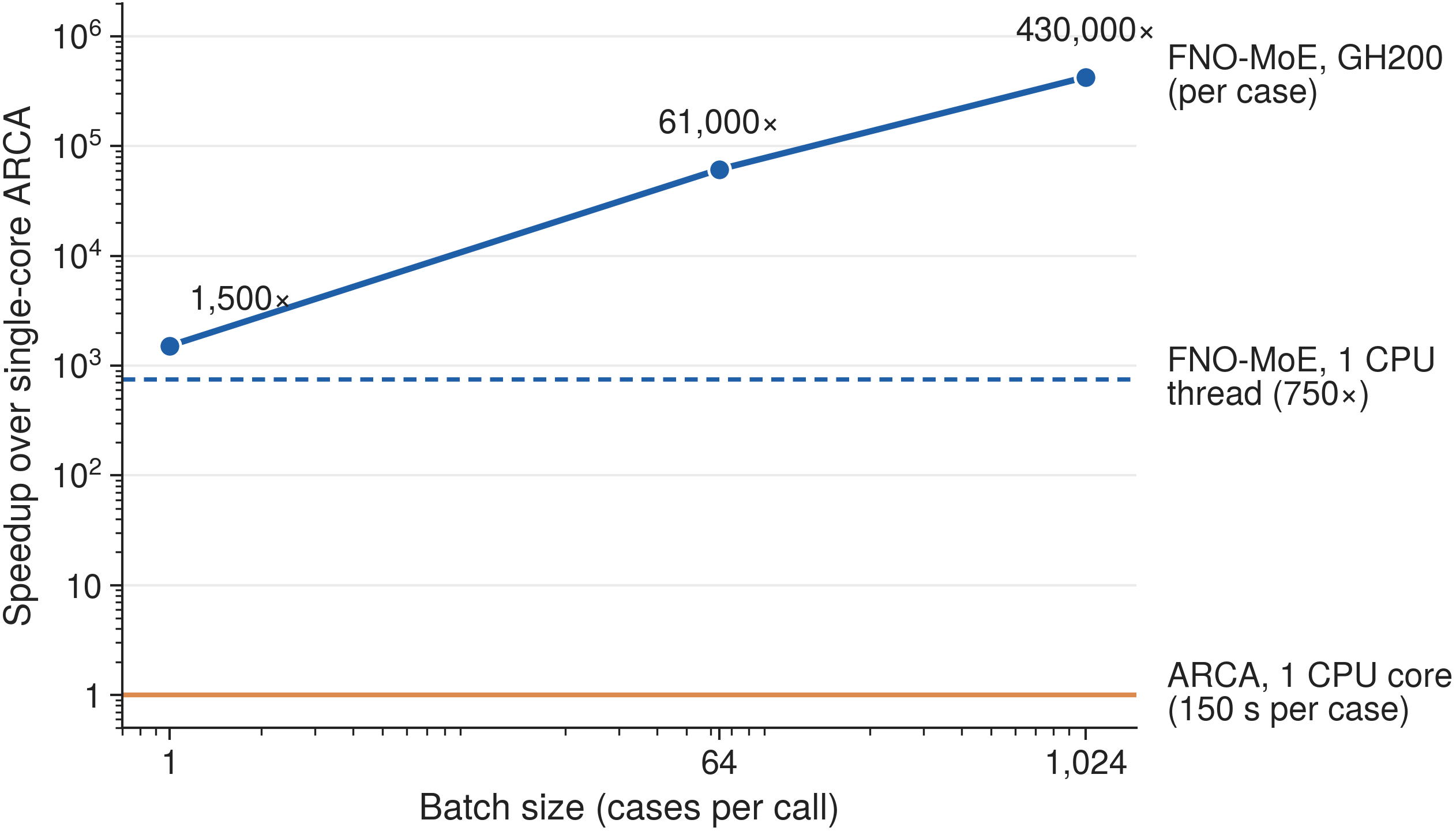}
\caption{\textbf{Batching amortizes GPU overhead.} Emulator speedup relative to the single-core ARCA reference (orange line, 150\,s per six-hour case) versus batch size on one GH200 GPU, using the amortized time per case; the dashed line is the emulator on one single-threaded CPU core. Coupled-model acceleration also depends on data movement, other processes and scheduling.}\label{sfig:timing}
\end{figure*}

\begin{figure*}[p]
\centering
\includegraphics[width=\textwidth,height=0.5\textheight,keepaspectratio]{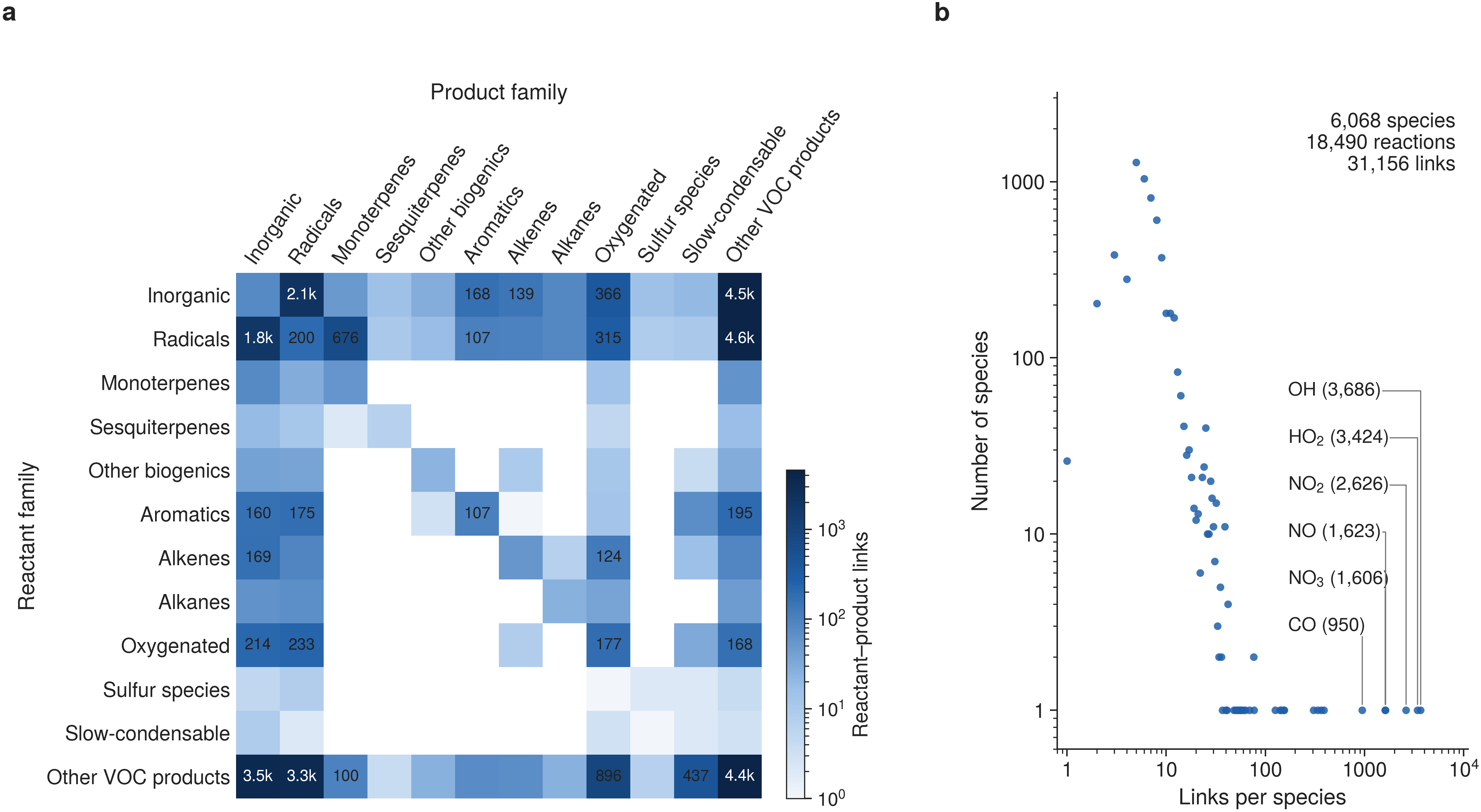}
\caption{\textbf{The configured mechanism is a dense network organized around a few inorganic and radical hubs.} Species--species graph in which a directed link joins a reactant to a product of the same reaction (6,068 species, 18,490 reactions, 31,156 reactant--product links). (a) Number of links from each reactant family (rows) to each product family (columns), on a logarithmic colour scale; counts of at least 100 are printed. (b) Distribution of the number of links per species; most species have fewer than 20 links, whereas OH (3,686), HO$_2$ (3,424), NO$_2$ (2,626), NO (1,623), NO$_3$ (1,606) and CO (950) connect to a large part of the mechanism. The emulator predicts the filtered retained state.}\label{sfig:network}
\end{figure*}

\begin{figure*}[p]
\centering
\includegraphics[width=\textwidth,height=0.5\textheight,keepaspectratio]{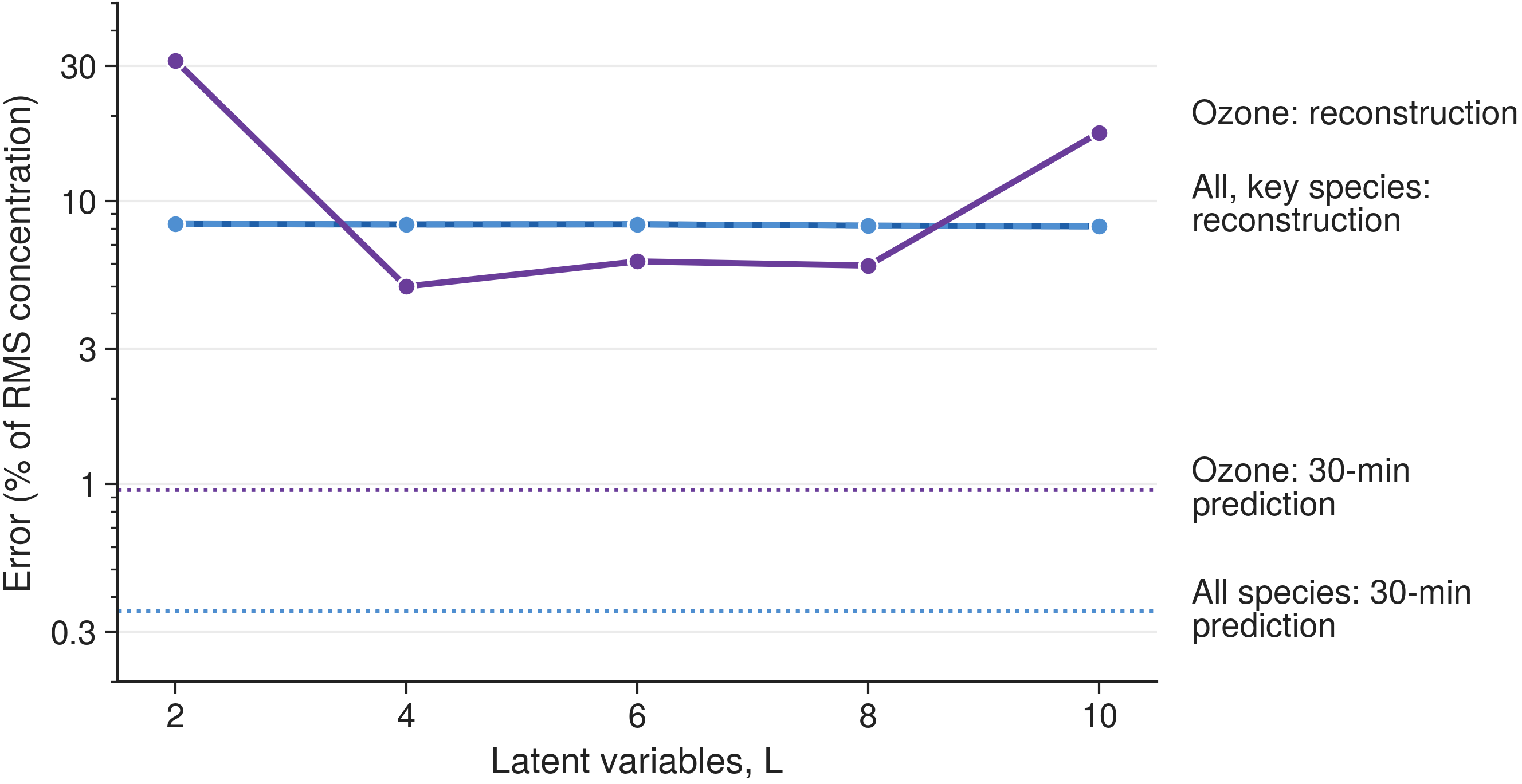}
\caption{\textbf{A low-dimensional linear code reconstructs the retained state with about 8\% error at every tested latent dimension.} Reconstruction error of the SIMADy-inspired linear autoencoder (\% of RMS concentration) versus the number of latent variables, for all species, key species and ozone. Dotted lines mark the 30-min prediction errors of 6h-NoStiff in the same units (0.36\% for all species, 0.95\% for ozone); the all-species and key-species curves coincide.}\label{sfig:ae}
\end{figure*}

\begin{figure*}[p]
\centering
\includegraphics[width=\textwidth,height=0.42\textheight,keepaspectratio]{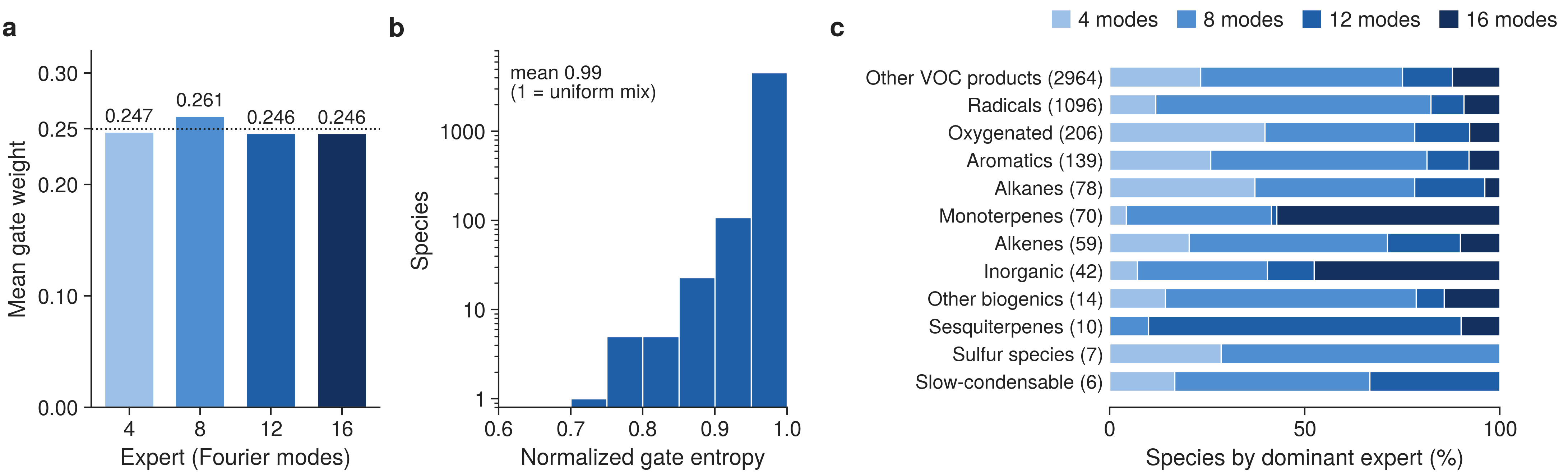}
\caption{\textbf{The expert gate is used jointly and does not collapse.} Experts are coloured light to dark by retained Fourier modes (4, 8, 12, 16), as in Fig.~\ref{fig:overview}. \textbf{a}, Mean softmax gate weight per expert; dotted line, uniform value 1/4. \textbf{b}, Distribution of per-species gate entropy (logarithmic count axis), normalized so that 0 is single-expert routing and 1 a uniform mix (mean 0.99). \textbf{c}, For each chemical family (species count in brackets), the share of species whose highest-weighted expert is each of the four.}\label{sfig:gate}
\end{figure*}

\begin{figure*}[p]
\centering
\includegraphics[width=\textwidth,height=0.5\textheight,keepaspectratio]{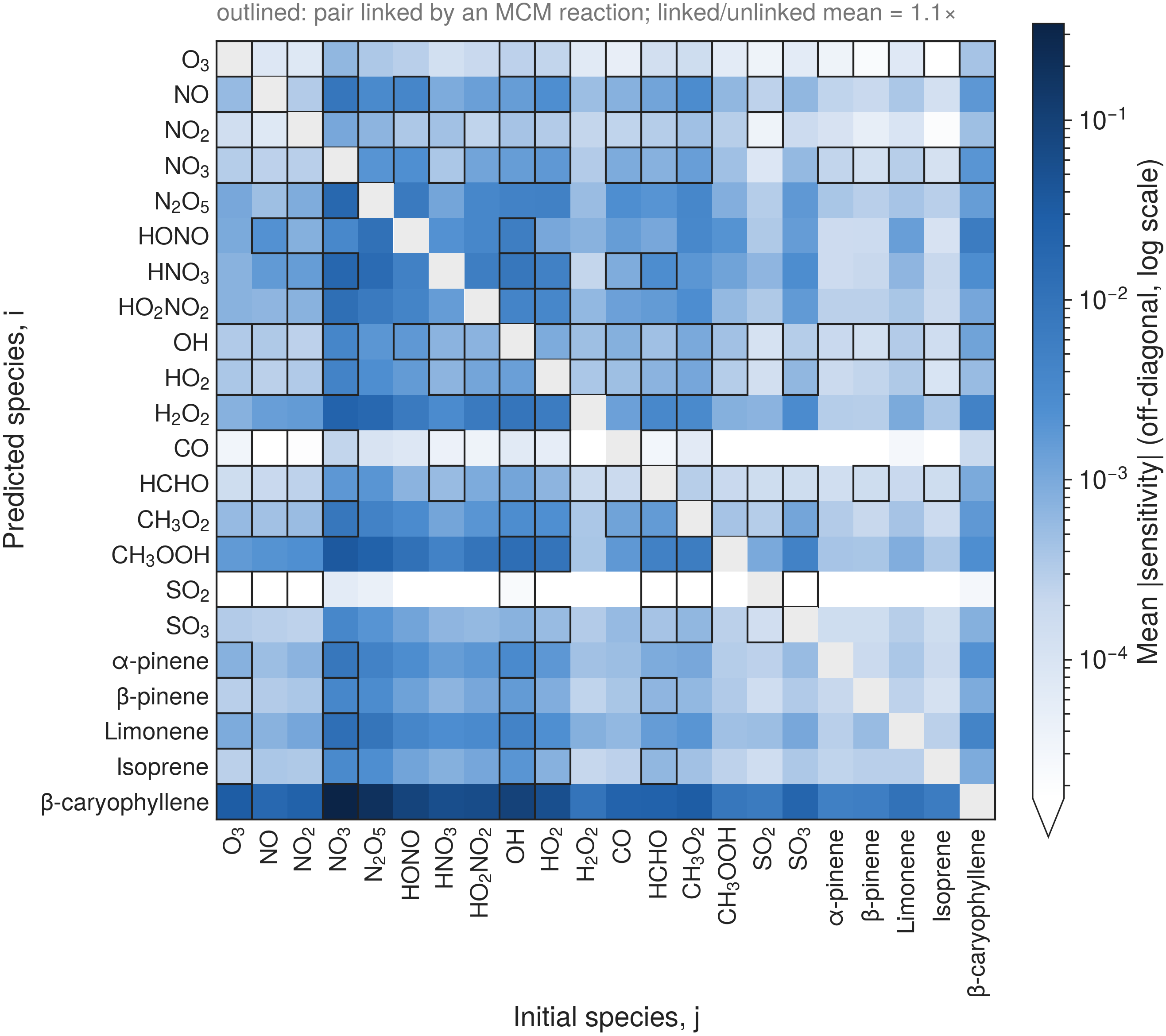}
\caption{\textbf{Finite-horizon sensitivities of the emulator.} Mean magnitude of the derivative of predicted species $i$ with respect to initial species $j$, $|\partial\widehat{X}_i/\partial x_{0,j}|$, averaged over 200 states for 22 inorganic, radical and biogenic species; outlined cells mark pairs directly linked by an MCM reaction, and the self-sensitivity diagonal (off scale) is hatched grey; colours use a logarithmic scale clipped at the 5th percentile of the off-diagonal values, so that structure outside the strongly sensitive $\beta$-caryophyllene row remains visible. Linked pairs are only 1.1 times as sensitive as unlinked pairs on average: sensitivities respond broadly to inorganic and radical inputs rather than to individual reaction links (Supplementary Note~\ref{snote:sens}).}\label{sfig:jacobian}
\end{figure*}

\begin{figure*}[p]
\centering
\includegraphics[width=\textwidth,height=0.5\textheight,keepaspectratio]{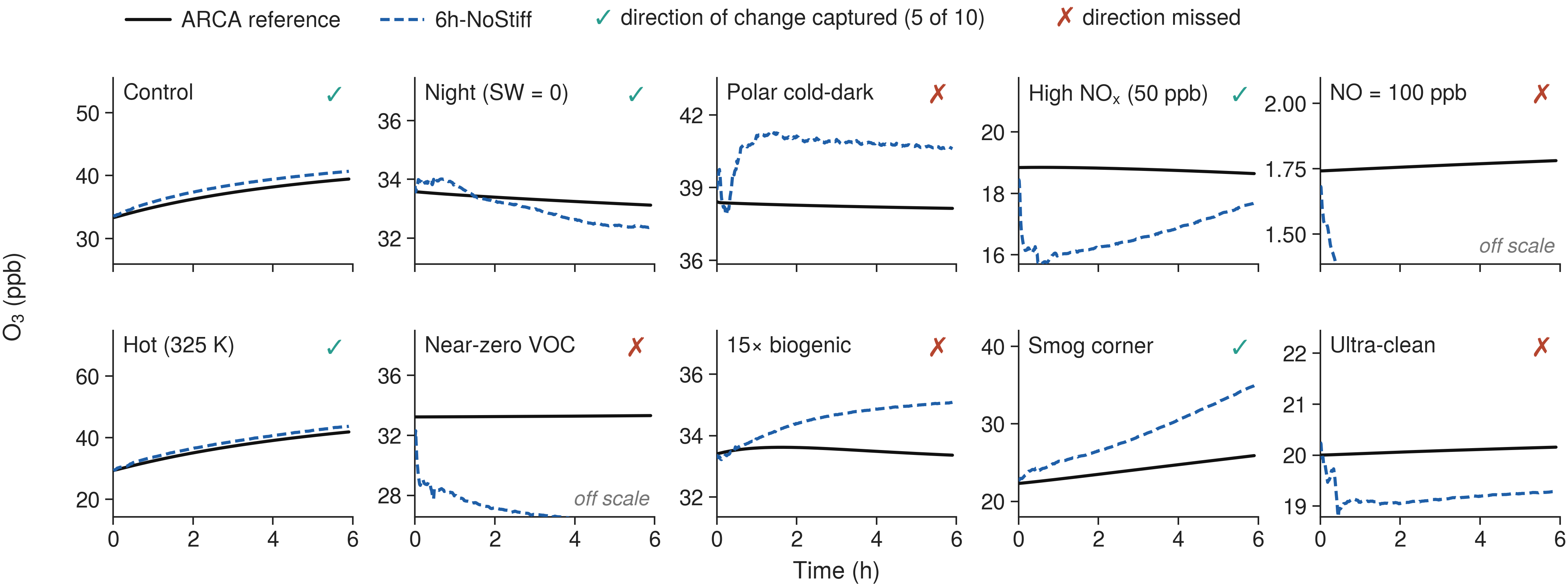}
\caption{\textbf{Ozone trajectories in the ten extreme scenarios.} Six-hour ozone from the reference (ARCA, solid) and the deployed 6h-NoStiff model (dashed) for each scenario, a realistic base state with one driver pushed to an extreme (Methods). Symbols mark whether the emulator captures the direction of the reference ozone change (5 of 10); panels marked ``off scale'' are clipped to the reference range. Vertical scales differ between panels. The direction diagram in Fig.~\ref{fig:generalization}c summarizes these trajectories.}\label{sfig:extreme}
\end{figure*}

\end{document}